\documentclass{article}

\usepackage{microtype}
\usepackage{graphicx}
\usepackage{booktabs} % for professional tables

\usepackage{subcaption}
\usepackage{float} 

\usepackage{hyperref}

\usepackage[accepted]{icml2023}

\usepackage{amsmath}
\usepackage{amssymb}
\usepackage{mathtools}
\usepackage{amsthm}

\usepackage[capitalize,noabbrev]{cleveref}

\theoremstyle{plain}

\theoremstyle{definition}

\theoremstyle{remark}

\usepackage[textsize=tiny]{todonotes}

\icmltitlerunning{Detecting and Restoring Non-Standard Hands in Stable Diffusion Generated Images}

\begin{document}

\twocolumn[
\icmltitle{Detecting and Restoring Non-Standard Hands in \\ Stable Diffusion Generated Images}



\icmlsetsymbol{equal}{*}

\begin{icmlauthorlist}
\icmlauthor{Yiqun Zhang}{equal,anu}
\icmlauthor{Zhenyue Qin}{equal,anu}
\icmlauthor{Yang Liu}{anu}
\icmlauthor{Dylan Campbell}{anu}
\end{icmlauthorlist}

\icmlaffiliation{anu}{Australian National University}

\icmlcorrespondingauthor{Zhenyue Qin}{kf.zy.qin@gmail.com}
\icmlcorrespondingauthor{Yiqun Zhang}{yiqun@admin.io}
\icmlcorrespondingauthor{Yang Liu}{lyf1082@gmail.com}
\icmlcorrespondingauthor{Dylan Campbell}{dylan.campbell@anu.edu.au}

\icmlkeywords{Machine Learning, ICML}

\vskip 0.3in
]



\printAffiliationsAndNotice{\icmlEqualContribution} % otherwise use the standard text.

% \begin{abstract}
% The Stable Diffusion model is a popular and effective model for image generation. But sometimes the image of the human hand it generates is not standard, such as a hand with less than or more than five fingers. Building upon the foundational HaGRID dataset, we curated our own dataset tailored to the specific challenges of non-standard hand representations. This research addresses this issue by introducing a comprehensive pipeline that not only detects these inaccuracies but also restores them to closely resemble real-world hand images, termed as standard hands. Our methodology incorporates a detection phase using a fine-tuned YOLO model, proficiently identifying and categorizing hand types across diverse datasets: images generated by Stable Diffusion, real photographs, and redrawn samples from the HaGRID dataset. Following detection, our multi-phased restoration process involves body pose estimation, control image generation, and subsequent inpainting processes, effectively transforming non-standard hand to their standard hand counterparts. The conducted experiments validate the robustness and efficacy of our approach, marking a significant advancement in enhancing the Stable Diffusion model's capabilities in hand image generation. For quick and easy use, we have encapsulated our methodology into an interactive web application. This platform empowers users to quick upload images and get immediate restoration feedback.
% \end{abstract}

\begin{abstract}
We introduce a pipeline to address anatomical inaccuracies in Stable Diffusion generated hand images. The initial step involves constructing a specialized dataset, focusing on hand anomalies, to train our models effectively. A finetuned detection model is pivotal for precise identification of these anomalies, ensuring targeted correction. Body pose estimation aids in understanding hand orientation and positioning, crucial for accurate anomaly correction. The integration of ControlNet and InstructPix2Pix facilitates sophisticated inpainting and pixel-level transformation, respectively. This dual approach allows for high-fidelity image adjustments. This comprehensive approach ensures the generation of images with anatomically accurate hands, closely resembling real-world appearances. Our experimental results demonstrate the pipeline's efficacy in enhancing hand image realism in Stable Diffusion outputs. We provide an online demo at \href{https://fixhand.yiqun.io}{fixhand.yiqun.io}.
\end{abstract}

\begin{figure*}[h]
    \centering
    \includegraphics[width=0.8\textwidth]{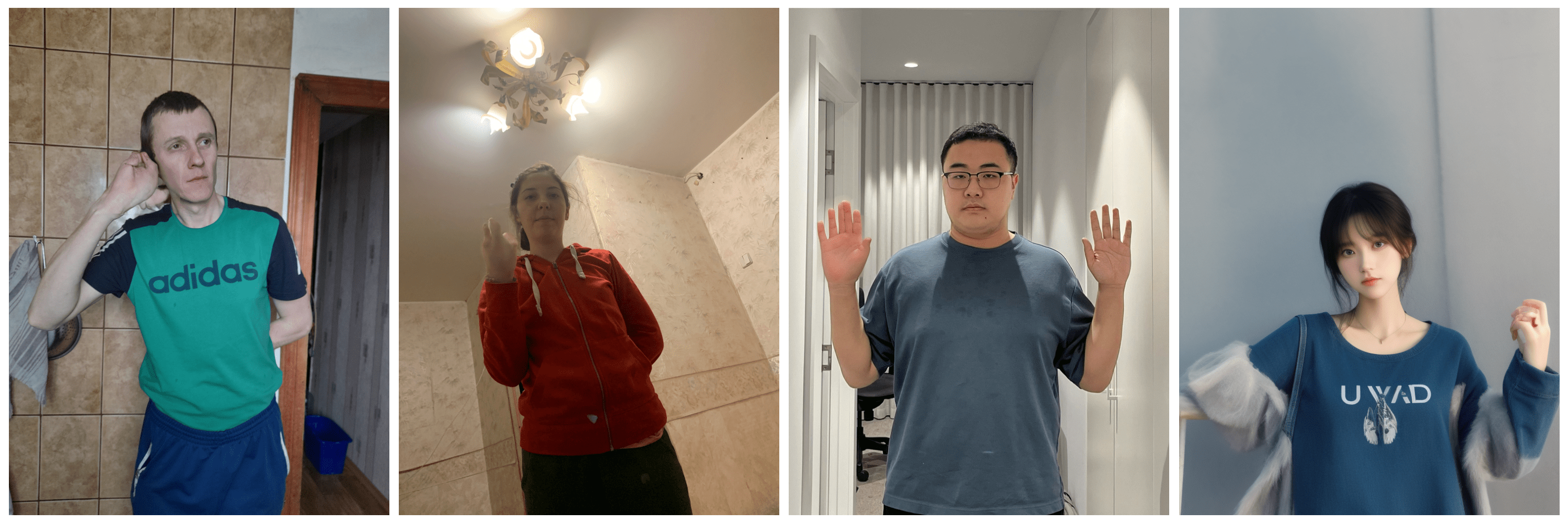}
    \caption{A compilation of images with non-standard hand anomalies, highlighting the varied manifestations of the issue.}
    \label{fig:intro-non-standard}
\end{figure*}

\begin{figure*}[!h]
    \centering
    \includegraphics[width=1\textwidth]{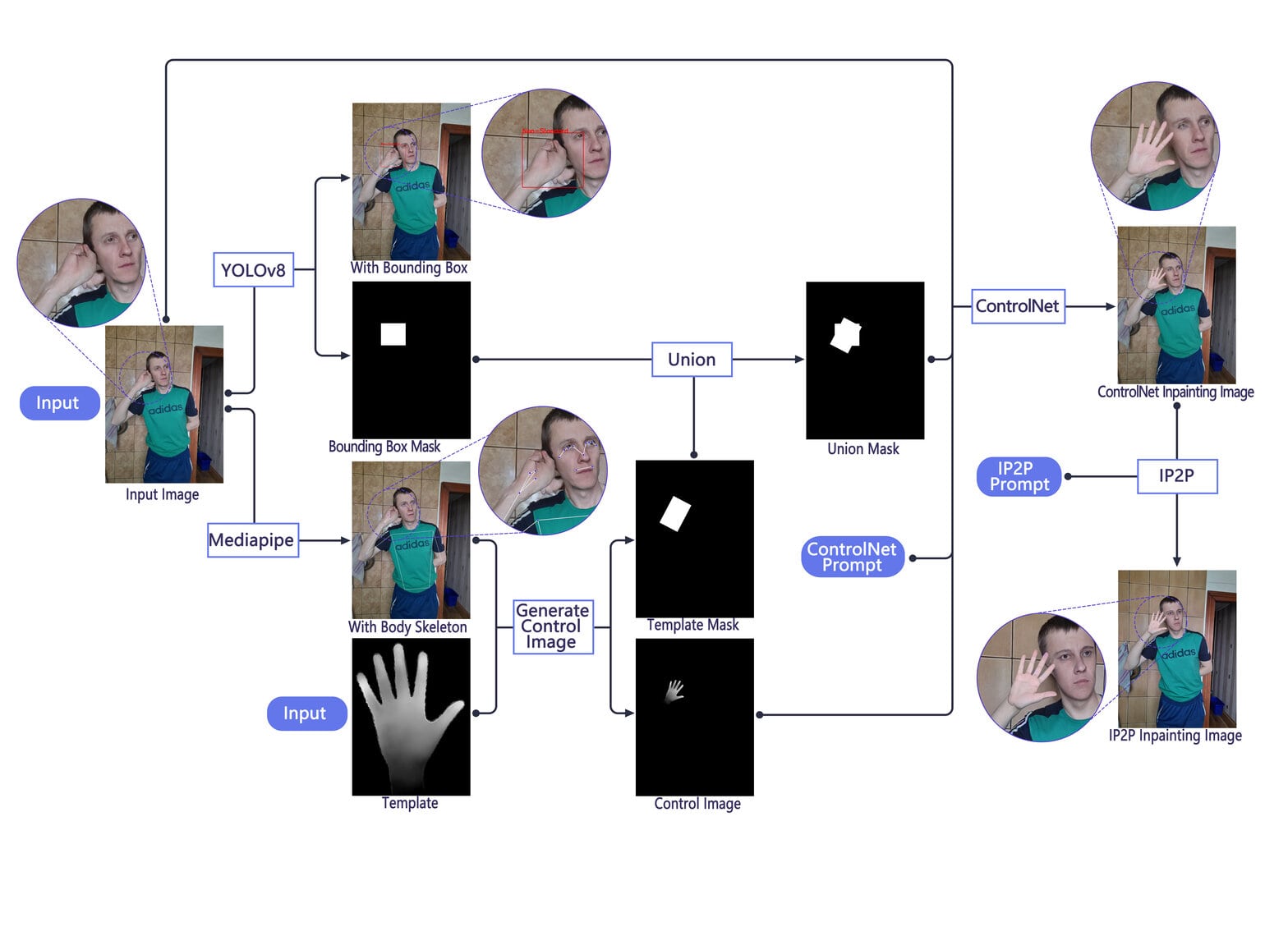}
    \vspace{-6mm}
    \caption{This flowchart outlines our proposed pipeline: Initially, an image with a non-standard hand as the input. We then employ YOLOv8 to delineate the non-standard hand using a bounding box, creating what we term the ``bounding box mask''. MediaPipe is utilized to compute the body skeleton. Based on this skeleton, a template is accurately positioned over the non-standard hand to create the ``control image''. The control image's bounding box and the bounding box mask are combined to generate the ``union mask''. Using this union mask, the control image, and a descriptive template prompt, we repair the area covered by the mask. Subsequently, IP2P and its associated prompt are used to refine the texture, resulting in the final output.}
    \label{fig:flowchart}
\end{figure*}

\section{Introduction} \label{sec:introduction}

% As the frontiers of artificial intelligence expand, the Stable Diffusion \cite{2022_ldm} model emerges as a formidable contender in the generation of human images. Its superior capability in crafting human likenesses, especially in real-time applications such as gaming or augmented reality, makes it an invaluable asset. However, we observe that the model frequently falters in generating hand images, leading to what we term as non-standard hand. As illustrated in \autoref{fig:intro-non-standard}, a non-standard hand can exhibit various anomalies such as missing fingers, an incorrect number of fingers, disproportionate size, or incorrect hand structure. Such inconsistencies, albeit subtle, can significantly compromise the overall realism and authenticity of the generated images.

Stable Diffusion~\cite{2022_ldm} has become increasingly prominent in generating human images. Its proficiency in creating realistic human representations, particularly for real-time applications like gaming or augmented reality, is noteworthy. However, a recurrent issue with this model is its tendency to produce inaccurate hand images, a problem we define as the "non-standard hand." Refer to \autoref{fig:intro-non-standard} for examples, where non-standard hands may display irregularities such as missing or extra fingers, disproportionate sizes, or structurally incorrect hands. These discrepancies, while minor, can greatly affect the perceived realism and authenticity of the images.

% The phenomenon of the \textit{uncanny valley}\cite{mori1970bukimi} suggests that when humanoid objects appear almost, but not exactly, like real human beings, it elicits feelings of eeriness and discomfort among human observers. In the context of Stable Diffusion's generation of hand images, non-standard hands risk plunging generated figures into this uncanny valley. Furthermore, in applications like augmented reality (AR), virtual reality (VR) and gaming, where immersion and realism are paramount, such irregularities can become particularly jarring for users. The disruption caused by these non-standard hands can hinder the user experience, limiting the full potential of these technologies. Ensuring accurate hand depiction is thus not only about aesthetics but also about the usability and user satisfaction in dynamic virtual environments.

The concept of the uncanny valley~\cite{mori1970bukimi} describes a sense of unease or discomfort when humanoid figures closely resemble humans but are not quite lifelike. This effect is relevant when considering the Stable Diffusion model's hand image generation, where images with "non-standard hands" may evoke the uncanny valley phenomenon. In fields like augmented reality (AR), virtual reality (VR), and gaming, where a seamless and realistic experience is crucial, such anomalies in hand images can be particularly unsettling for users. Non-standard hands may disrupt user engagement, detracting from the effectiveness of AR, VR, and gaming applications. Accurate hand representation is, therefore, essential not just for visual appeal, but also for the functionality and overall user satisfaction in interactive virtual settings.

% Besides Stable Diffusion, other image generative models such as Variational Autoencoders (VAEs) \cite{kingma2013auto} and Generative Adversarial Networks (GANs) \cite{goodfellow2014generative} also exist. However, compared to the adversarial training required by GANs or the variational posterior needed by VAEs, the predominant advantage of the Diffusion model lies in its simplicity of training. Leveraging the architecture of UNet \cite{ronneberger2015u} from the domain of image segmentation, the Stable Diffusion model boasts a stable training loss and exceptional performance. We perceive Stable Diffusion to be of greater practical value. This inherent strength of Stable Diffusion has driven our focus towards addressing the non-standard hand anomalies specifically within Stable Diffusion.

Other models for creating images, like Variational Autoencoders (VAEs) \cite{kingma2013auto} and Generative Adversarial Networks (GANs) \cite{goodfellow2014generative}, exist besides Stable Diffusion. But the easier training of the Diffusion model is a big plus over GANs, which need adversarial training, and VAEs, which need a variational posterior. The Stable Diffusion uses UNet architecture \cite{ronneberger2015u} from image segmentation and shows stable loss during training and very good performance. We think Stable Diffusion is more useful for practical situations. That's why we focus on solving the problem of non-standard hands in Stable Diffusion.

% Despite the evident necessity to resolve this challenge, there's a conspicuous scarcity of comprehensive research in this realm. To bridge this gap, our work introduces a systematic methodology that first detects the non-standard hand and then restores them to resemble real-world hand images, which we denote as standard hand. Our proposed pipeline seamlessly integrates detection and restoration. The process kicks off with the detection phase, which is engineered to identify both standard and non-standard hand types in images using bounding boxes. Harnessing a dedicated dataset and leveraging the prowess of the YOLOv8 model \cite{yolov8}, our method achieves commendable precision in this phase, example of detection phase is illustrated in \autoref{fig:intro-detection}.

\begin{figure*}[!tb]
    \centering
	\subfloat[Image with both standard and non-standard-hand.]{
        \label{fig:intro-detection-1}
        \includegraphics[height=5.6cm]{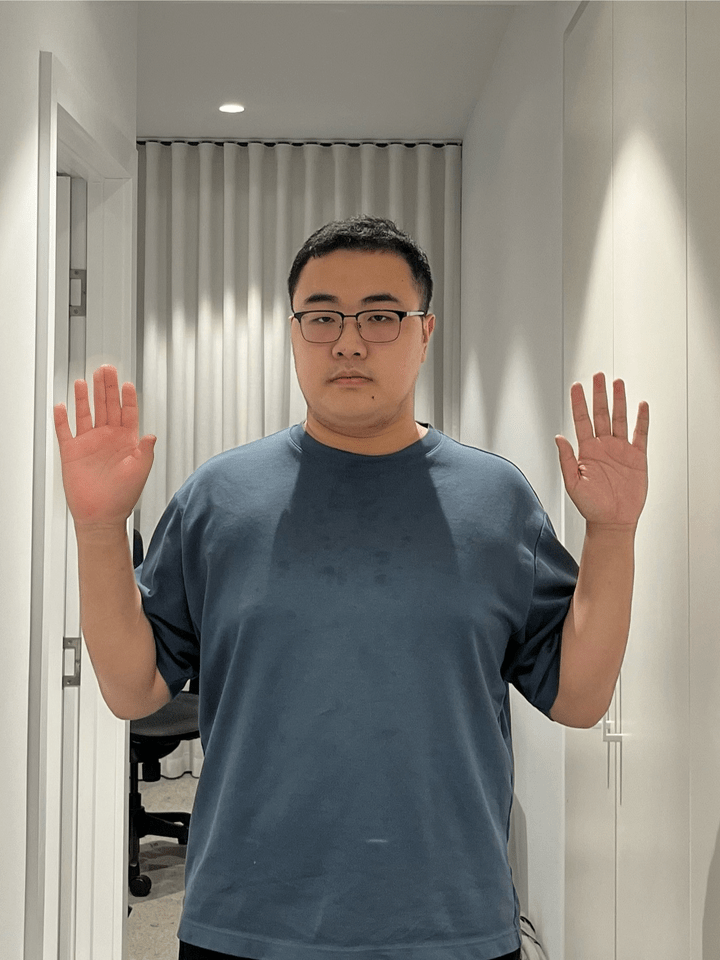}
    }
	\subfloat[Result of detection.]{
        \label{fig:intro-detection-2}
        \includegraphics[height=5.6cm]{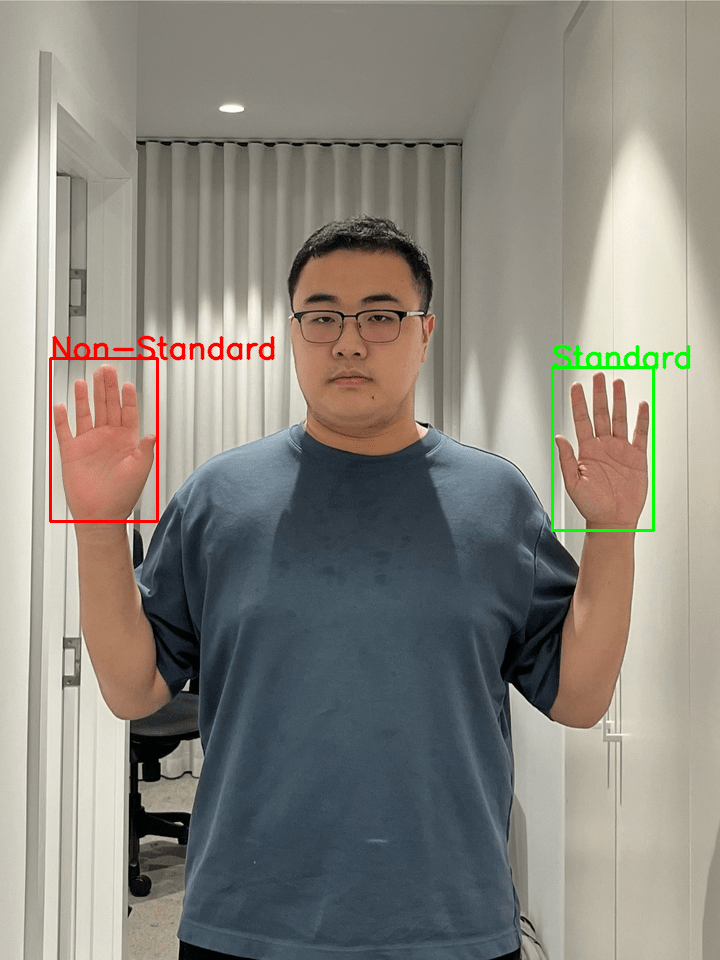}
    }
	\subfloat[Mask of only non-standard hand.]{
        \label{fig:intro-detection-3}
        \includegraphics[height=5.6cm]{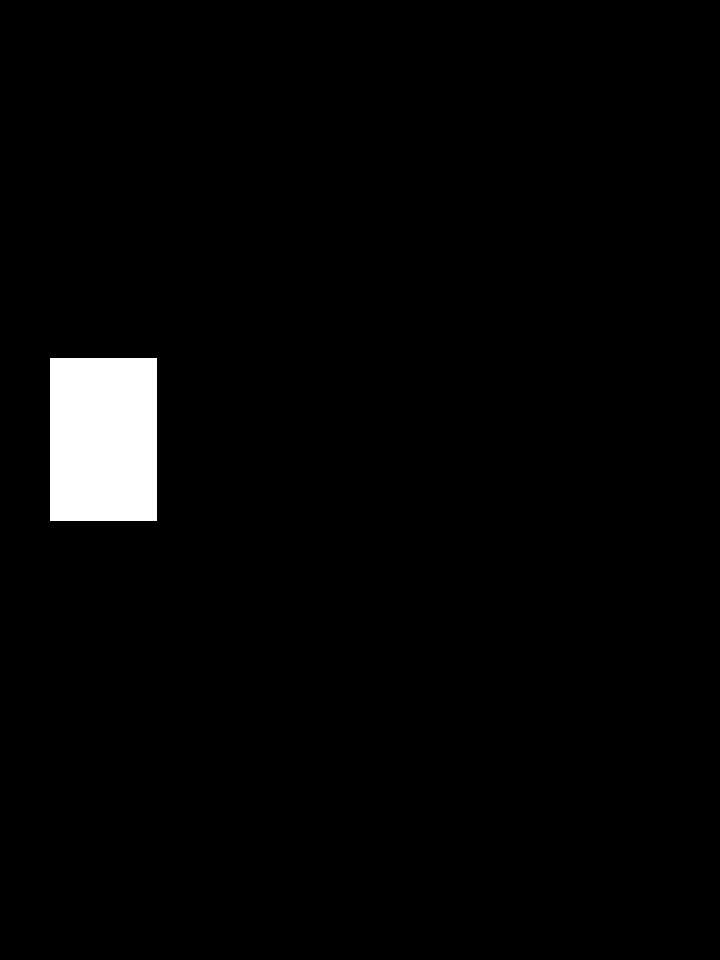}
    }
    \caption{Examples of detection. Image is an author's photo. }
    \label{fig:intro-detection}
\end{figure*}

Despite the evident necessity to resolve this challenge, there's a conspicuous scarcity of comprehensive research in this area. Our contribution is a structured method that initially identifies 'non-standard' hand depictions and subsequently corrects them to match the appearance of actual human hands, referred to here as ``standard hands''. We have devised an integrated pipeline for detection and correction. The initial phase employs bounding boxes to distinguish between ``standard'' and ``non-standard'' hands within images. Utilizing a specialized dataset and the advanced capabilities of the YOLOv8 algorithm \cite{yolov8}, our approach attains notable accuracy in detection, with \autoref{fig:intro-detection} demonstrating an instance of the detection phase.

% With the anomalies detected, our approach transitions to the restoration phase. Here, our pipeline meticulously follows a sequence of steps: \textit{Body Pose Estimation}, which leverages Google's MediaPipe \cite{mediapipe} to establish hand orientation; \textit{Control Image Generation}, a key step that guides the subsequent restoration by producing instructive images for the Stable Diffusion model; \textit{ControlNet Inpainting}, wherein the image undergoes initial restoration aided by the Control Image; and, finally, \textit{IP2P Inpainting}, a finishing touch to impart the generated hand images with life-like textures and authentic appearances. The pipeline overview is shown in \autoref{fig:intro-pipeline}.

\begin{figure*}[!tb]
    \centering
	\subfloat[Images with non-standard-hand.]{
        \includegraphics[height=5.6cm]{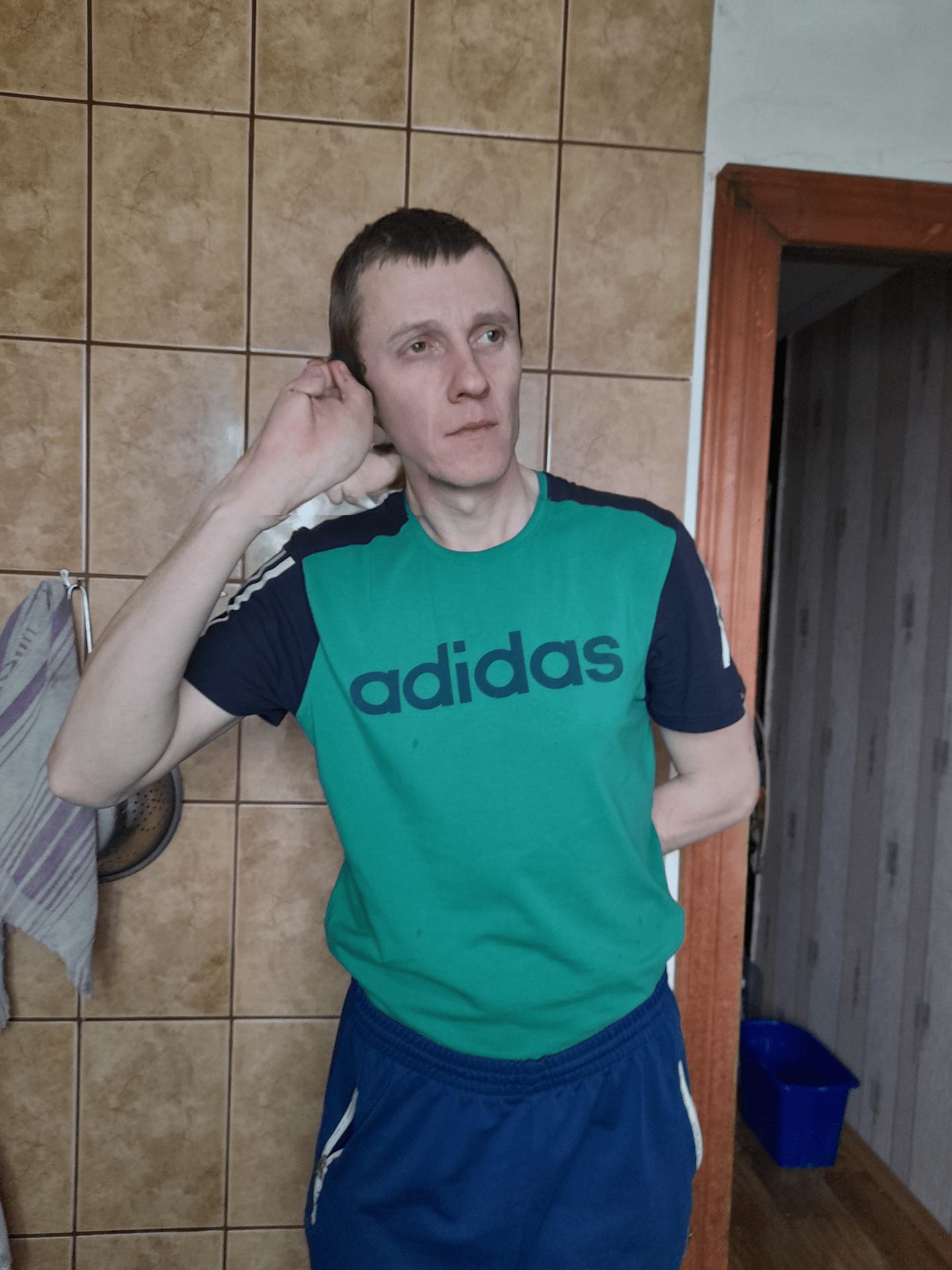}
    }
	\subfloat[Zoom in of non-standard-hand.]{
        \includegraphics[height=5.6cm]{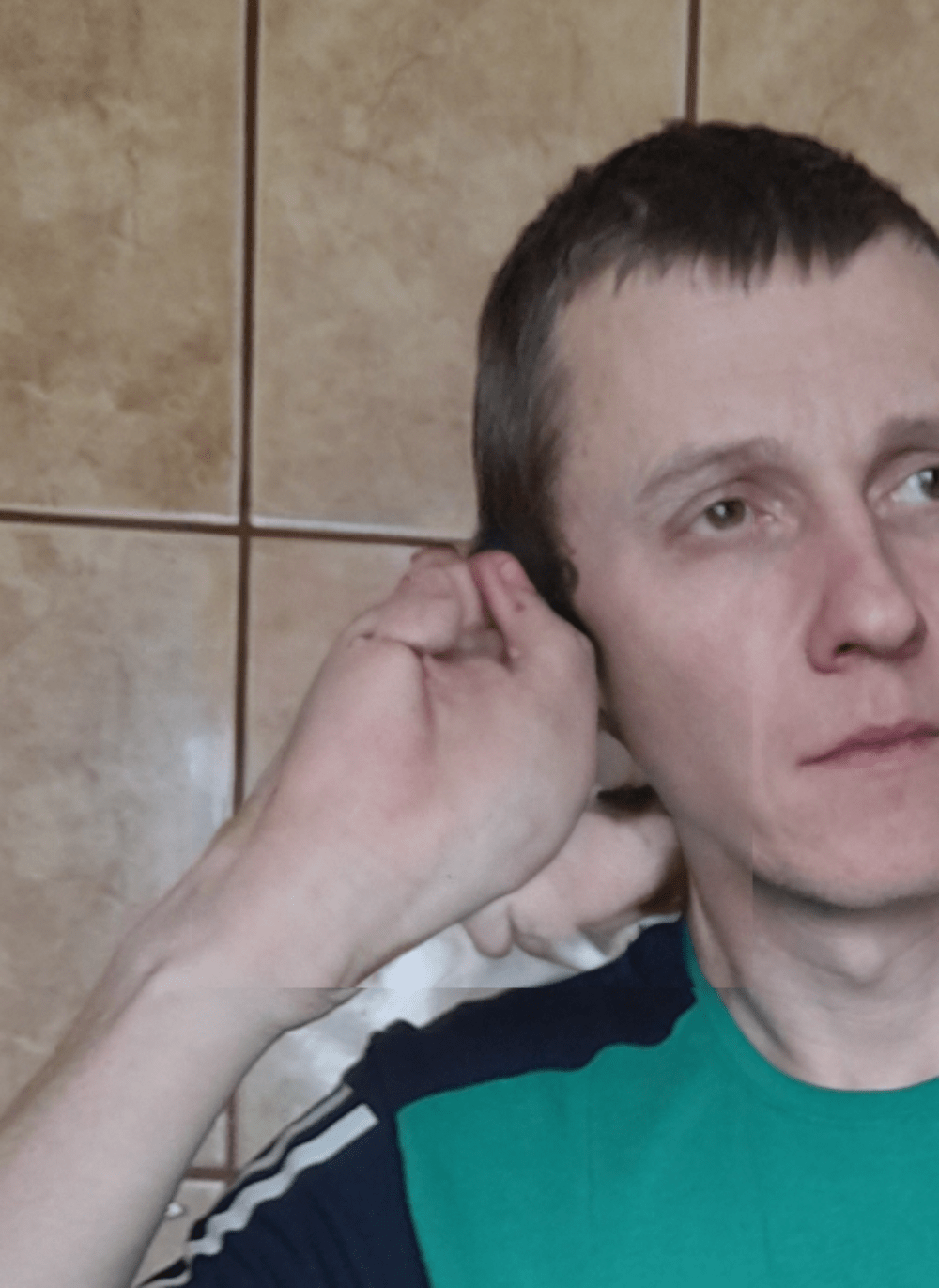}
    }
	\subfloat[Result of non-standard hand detection.]{
        \includegraphics[height=5.6cm]{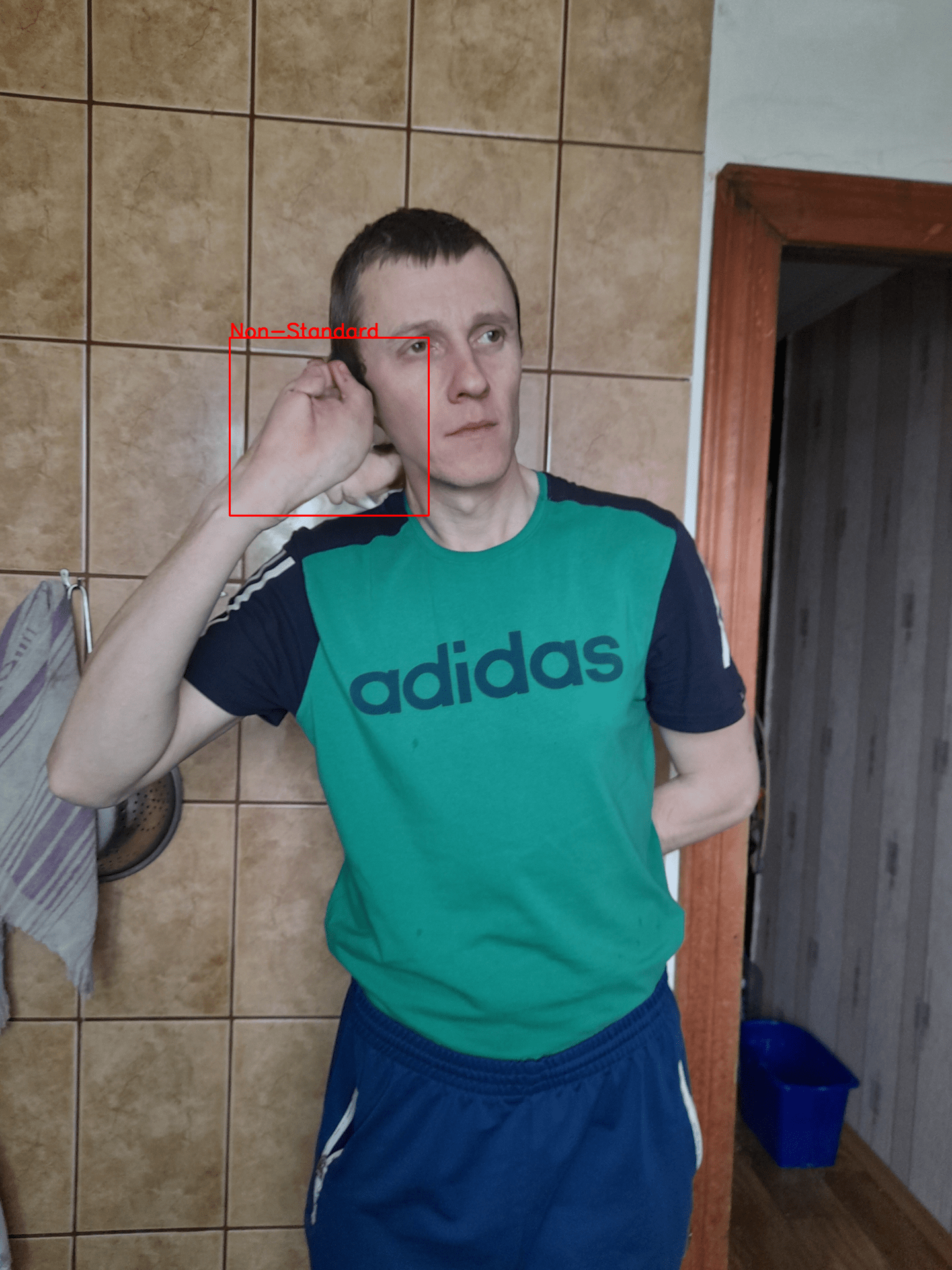}
    }
    \\
    \subfloat[Control image visualization.]{
        \includegraphics[height=5.6cm]{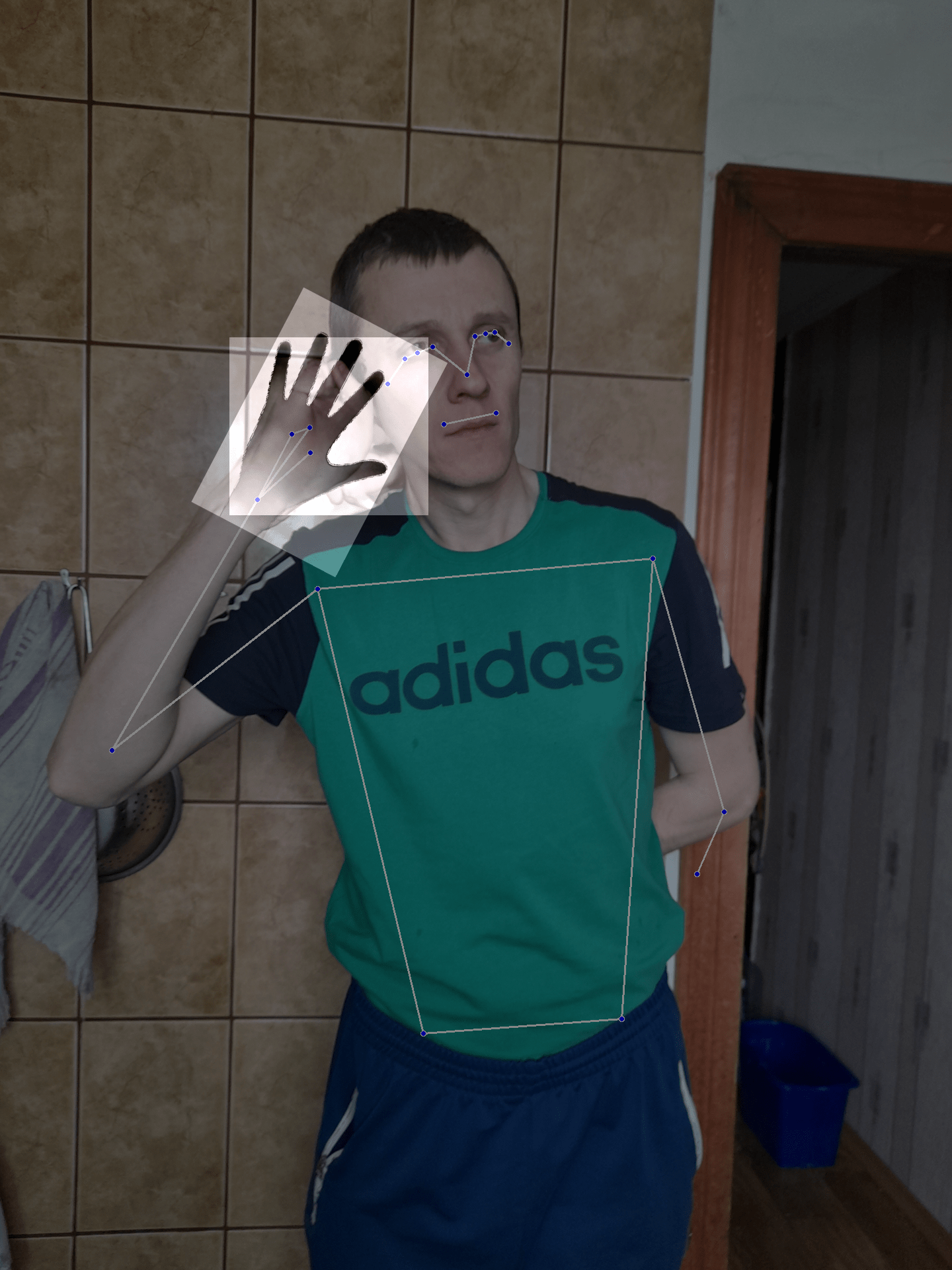}
    }
	\subfloat[Zoom in of control image visualization.]{
        \includegraphics[height=5.6cm]{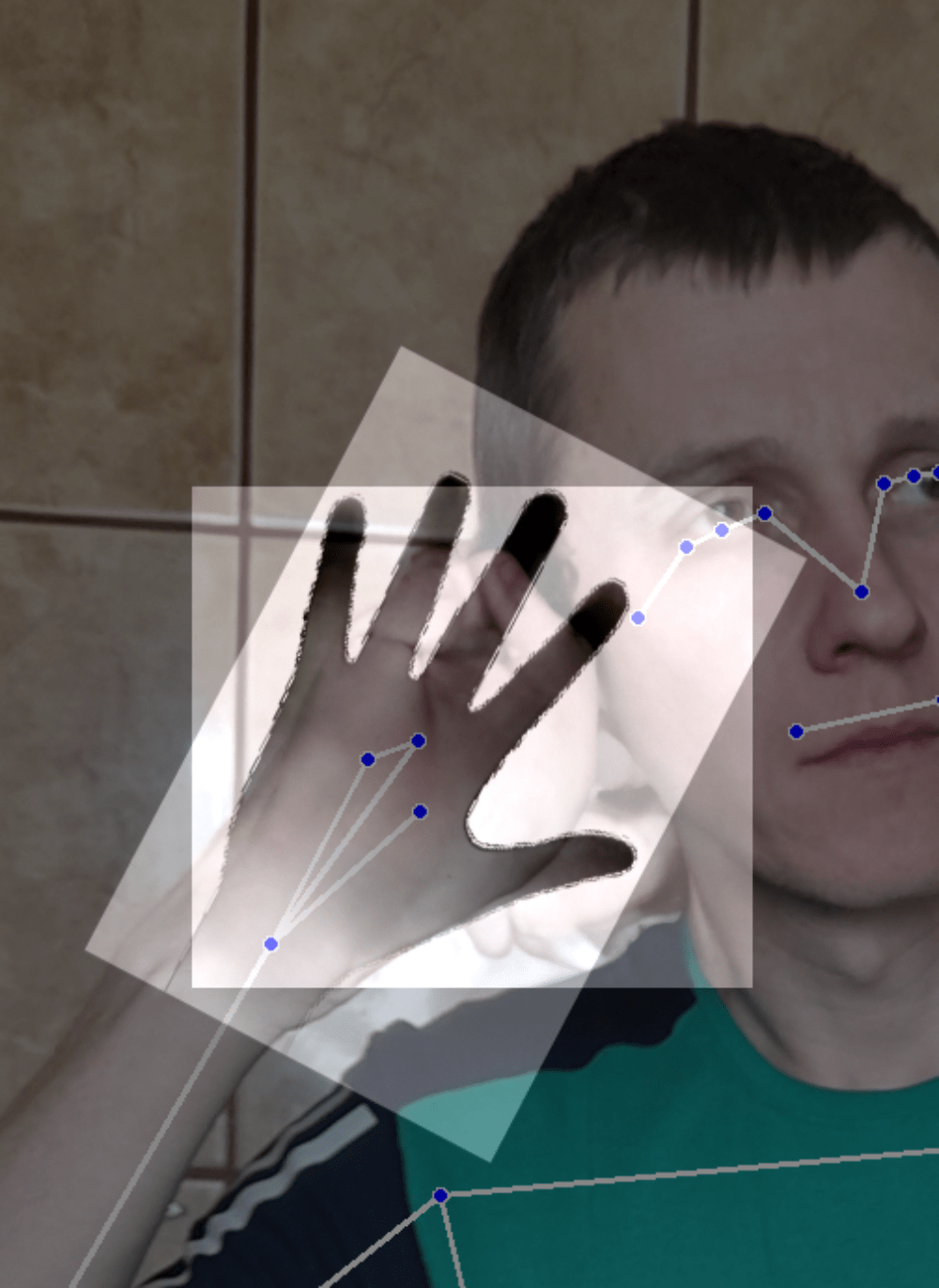}
    }
	\subfloat[ControlNet inpainting.]{
        \includegraphics[height=5.6cm]{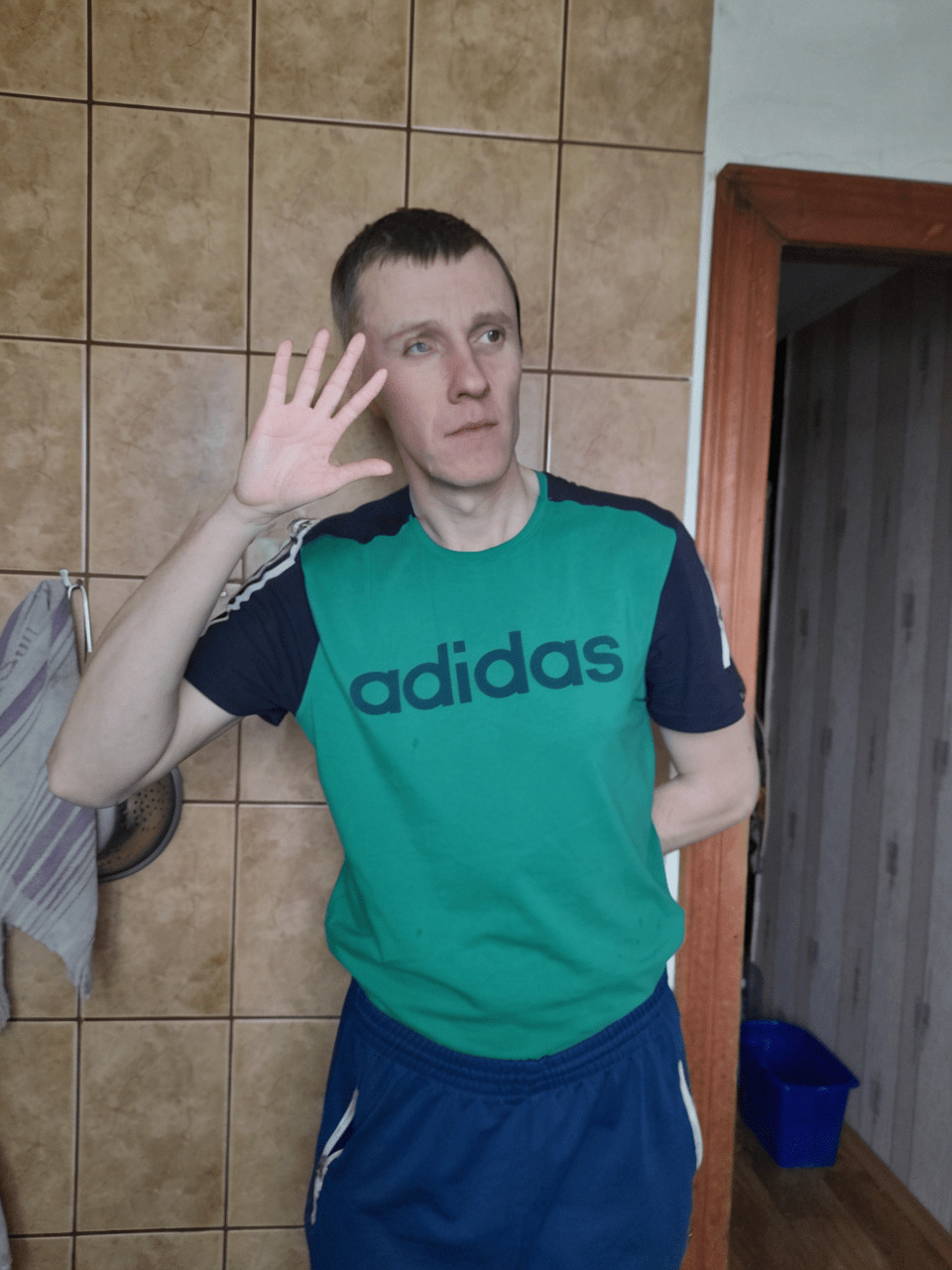}
    }
    \\
    \subfloat[Zoom in of controlNet inpainting.]{
        \includegraphics[height=5.6cm]{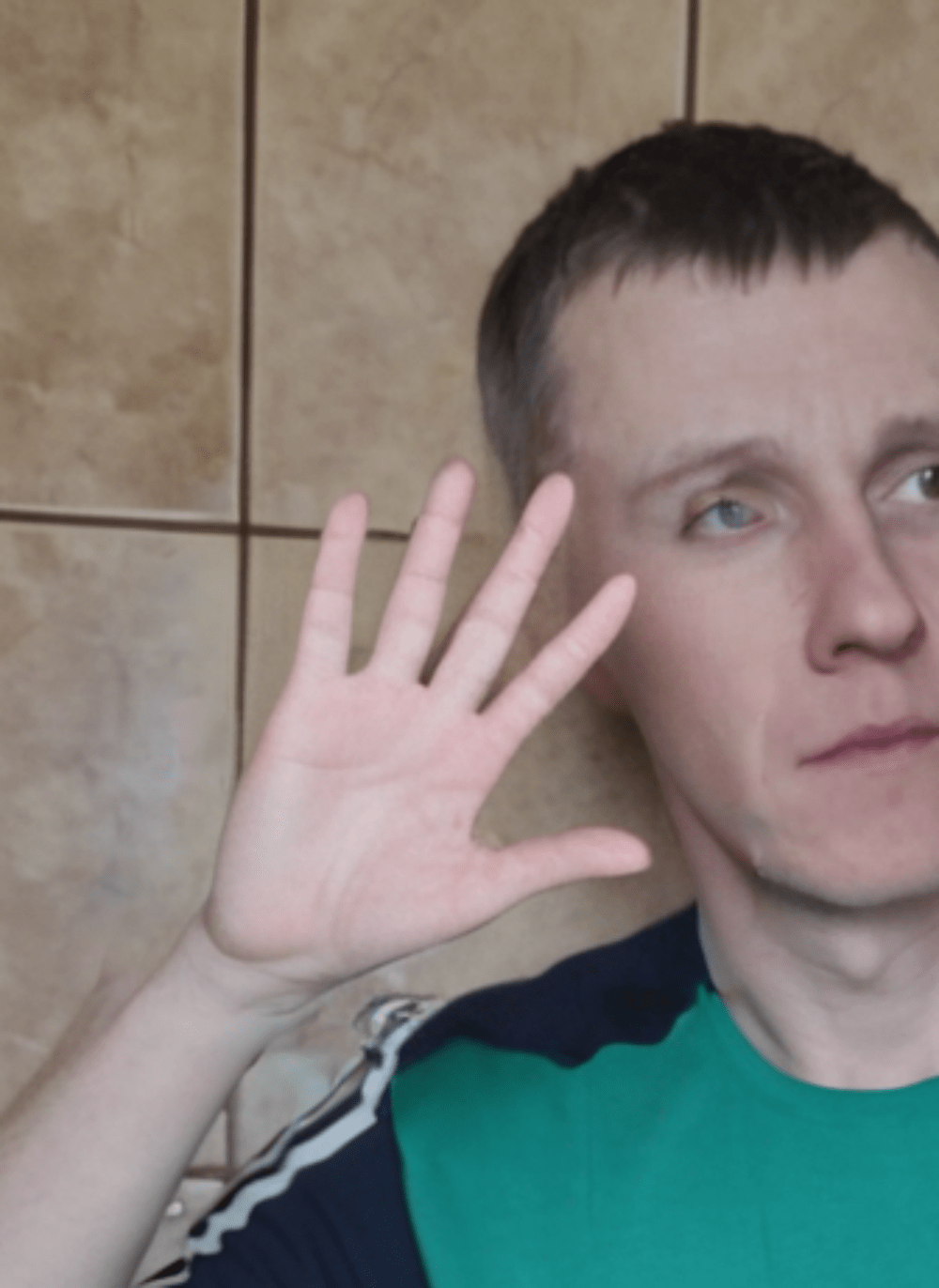}
    }
	\subfloat[IP2P inpainting.]{
        \includegraphics[height=5.6cm]{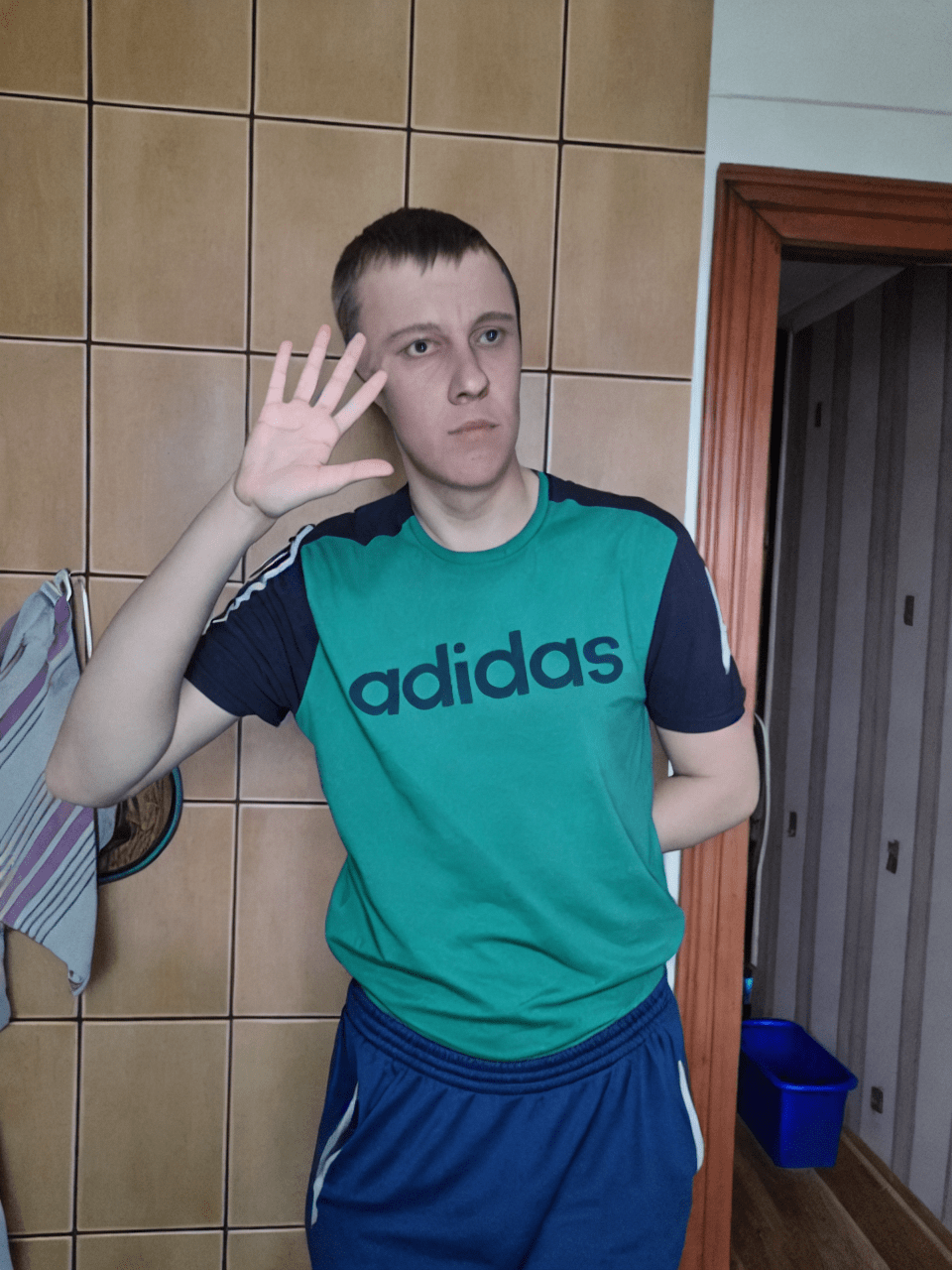}
    }
	\subfloat[Zoom in of IP2P inpainting.]{
        \includegraphics[height=5.6cm]{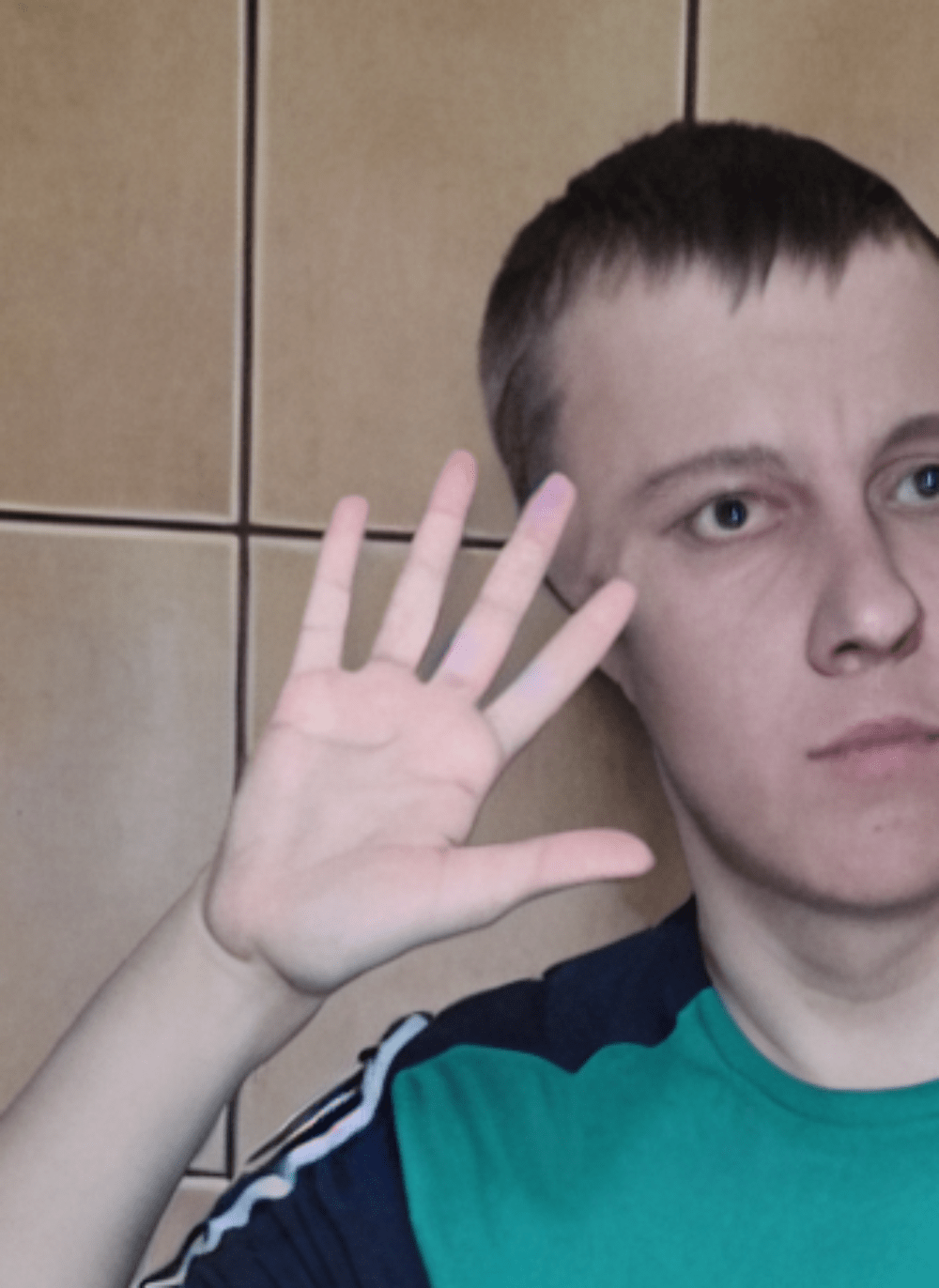}
    }
    \caption{Pipeline overview.}
    \label{fig:intro-pipeline}
\end{figure*}

Upon identifying anomalies, our system progresses to the restoration phase. The process involves a set of defined operations: \textit{Body Pose Estimation}, utilizing Google's MediaPipe \cite{mediapipe} to determine the hand's position and movement; \textit{Control Image Generation}, which provides the Stable Diffusion model with directive images for better restoration outcomes; \textit{ControlNet Inpainting}, offering an initial refinement based on the Control Image; and ultimately, \textit{IP2P Inpainting}, which enhances the images with realistic textures and accurate details. An illustration of this pipeline is depicted in \autoref{fig:intro-pipeline}.

% After our experiments in \autoref{sec:experiment}, our pipeline has achieved good results. Our fine-tuned YOLOv8 model impressively distinguishes between non-standard hand and standard hand. Moreover, during the restoration phase, our methodology demonstrably and gracefully transforms the non-standard hand to closely match the standard variant.

% In order to bridge the gap between advanced research methodologies and end-user accessibility, we have developed an interactive web application that encapsulates our novel hand-restoration approach. This user-friendly platform allows users to swiftly upload images containing non-standard hands and, within moments, receive immediate restoration feedback. Our intention behind this design is to ensure that our sophisticated methods can be easily leveraged, providing both enthusiasts and professionals a firsthand experience of the transformative power of our technology.

In this report, we present a solution that improves the image quality produced by Stable Diffusion, particularly the depiction of hands. Our finetuned YOLOv8 model effectively differentiates between non-standard and standard hand representations. Furthermore, in the restoration phase, our pipeline converts non-standard hands to more accurately resemble their standard counterparts.

We summarize our contributions as follows: 

% \begin{enumerate}
%     \item We curated a new dataset by leveraging the HaGRID dataset \cite{hagrid}. The dataset has a lot of labeled standard and non-standard hands, preparing for the next standard and non-standard hand to be detected.
    
%     \item With the objective of achieving precise detection and classification capabilities, the YOLOv8 model was fine-tuned. This enhanced version exhibits a remarkable prowess in distinguishing between standard and non-standard hand images.
    
%     \item Our work introduces a dedicated pipeline aimed at non-standard hand restoration. This pipeline, uniquely crafted, ensures meticulous attention to the restoration process, consequently refining the final image quality.
    
%     \item The efficacy of our restoration methodology has been solidified through rigorous experiments. Our results corroborate that not only is our approach effective within the Stable Diffusion generated images, but it also exhibits commendable restoration capabilities on images with diverse styles.
    
% \end{enumerate}

\begin{enumerate}
    \item We create a pipeline to rectify anatomical inaccuracies in hand images Our results demonstrate the effectiveness in producing anatomically accurate and realistic hand images in outputs from Stable Diffusion. 
    \item We finetune a detection model to locate and classify standard and nonstandard hands, and fine-tuned an InstructPix2Pix model to make high-fidelity adjustments to the images. We will make these models available. 
    \item We create a dataset featuring a diverse collection of hand images, both standard and nonstandard, to facilitate comprehensive model training. We plan to release this dataset publicly in the future. 
    \item Demo of this pipeline is available: \href{fixhand.yiqun.io}{fixhand.yiqun.io}.
\end{enumerate}

\section{Method}\label{sec:method}

% \subsection{Overview}

% In the images generated by the Stable Diffusion, there exists a notable challenge: the frequent representation of hands that deviate from conventional appearance, termed as non-standard hand. Our methodology is structured to detect these anomalies and subsequently restore them to closely match real-world hand images, which we refer to standard hand.

% Our approach can be broadly divided into two primary segments: detection and restoration. In the detection phase, we focus on identifying the presence of both standard and non-standard hands in images, marking them with bounding boxes.. This is achieved through the creation of a dedicated dataset and the fine-tuned detection model. Following the detection, the Restoring phase is engaged. Detection and restoration are combined into a pipeline, the steps are \textit{Non-standard Hand Detection}, \textit{Body Pose Estimation}, \textit{Control Image Generation}, \textit{ControlNet Inpainting} and \textit{IP2P Inpainting}, are illustrated in Fig1. Through the steps, our approach can restore non-standard hand. The subsequent sections delve deeper into the intricacies of each segment. The flowchart is displayed in \autoref{fig:flowchart}.

Stable Diffusion occasionally generates images with atypical hands, defined as non-standard hands. Our method is to identify these variations and then adjust them to resemble real-world hands, defined as standard hands.

Our strategy consists of two main parts: detection and restoration. The detection stage involves pinpointing both standard and non-standard hands in images, highlighting them with bounding boxes. This step relies on a specialized dataset and a finely-tuned detection model. After detection, we proceed to the restoration phase. Our pipeline integrates these phases, including steps like \textit{Non-standard Hand Detection}, \textit{Body Pose Estimation}, \textit{Control Image Generation}, \textit{ControlNet Inpainting}, and \textit{IP2P Inpainting}, as shown in~\autoref{fig:intro-non-standard}. These steps collectively enable the correction of non-standard hands. The following sections provide detailed insights into each part of the process. The entire workflow is depicted in \autoref{fig:flowchart}.

% \begin{figure*}[!bt]
%     \centering
%     \includegraphics[width=1\textwidth]{images/flowchart.jpg}
%     \caption{This flowchart outlines our proposed pipeline: Initially, an image with a non-standard hand as the input. We then employ YOLOv8 to delineate the non-standard hand using a bounding box, creating what we term the ``bounding box mask''. MediaPipe is utilized to compute the body skeleton. Based on this skeleton, a template is accurately positioned over the non-standard hand to create the ``control image''. The control image's bounding box and the bounding box mask are combined to generate the ``union mask''. Using this union mask, the control image, and a descriptive template prompt, we repair the area covered by the mask. Subsequently, IP2P and its associated prompt are used to refine the texture, resulting in the final output.}
%     \label{fig:flowchart}
% \end{figure*}

\subsection{Non-Standard Hand Detection}

Non-standard hand detection phase is designed to locate the bounding boxes of all hands present within an image. Moreover, it categorizes these hands into two distinct categories: non-standard hand and standard hand.

\textbf{Non-Standard Hand Dataset} \label{section:method_detecting_database}

% In order to achieve a precise localization and classification of hands within images, we need to create a dataset. This dataset should contain many images with hands. Notably, each hand in these images must be annotated, not only with its bounding box but also with its classification label denoting whether it is a non-standard hand or a standard hand. The dataset forms the bedrock of our methodology. 

For accurate localization and categorization of hands in images, constructing a dedicated dataset is essential. This dataset should encompass a wide range of images featuring hands. Crucially, every hand in these images requires annotation. This includes both the bounding box around the hand and a classification label indicating whether it is a non-standard or standard hand. Such a dataset is fundamental to the effectiveness of our approach.

\begin{figure*}[!bt]
    \centering
    \includegraphics[height=11.2cm]{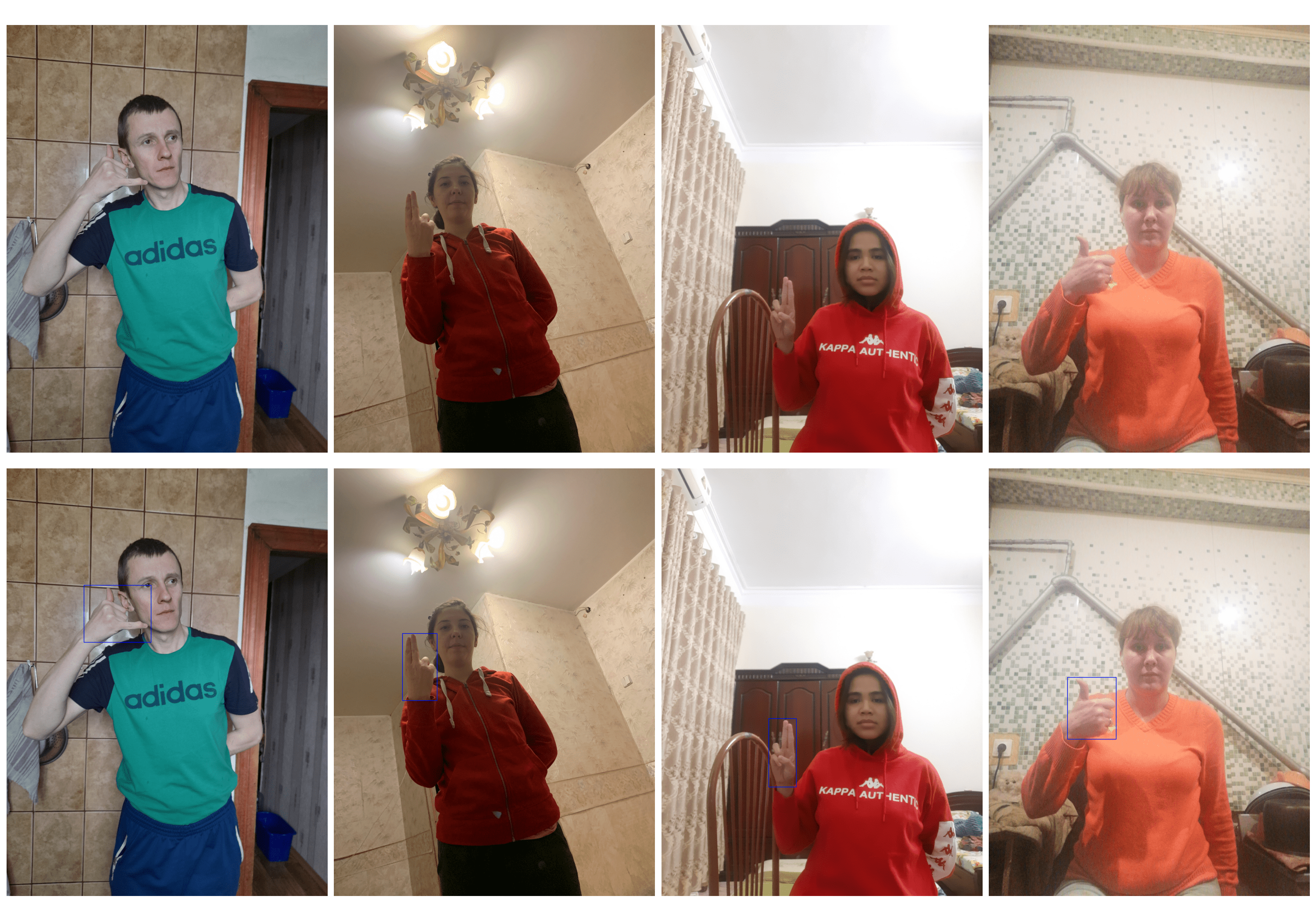}
    \caption{HaGRID dataset samples and samples with bounding boxes.}
    \label{fig:hagrid}
\end{figure*}

We begin with HaGRID (HAnd Gesture Recognition Image Dataset)~\cite{hagrid} as our foundational dataset. HaGRID comprises 552,992 RGB images, each containing hands in a variety of positions. Each image is annotated with bounding boxes to indicate hand locations. The dataset represents a wide array of 18 hand gestures, performed by a diverse group of 34,730 individuals, aged 18 to 65. Some images depict one hand, while others show both. The photos were taken indoors under different lighting conditions, providing a rich assortment of visual details. This diversity grants our dataset extensive coverage and high levels of generalizability, enhancing our model's capability to process hands in a wide range of appearances. Illustrations from HaGRID, including examples of bounding box annotations, are shown in \autoref{fig:hagrid}.

\begin{figure*}[!bt]
    \centering
    \includegraphics[height=5.6cm]{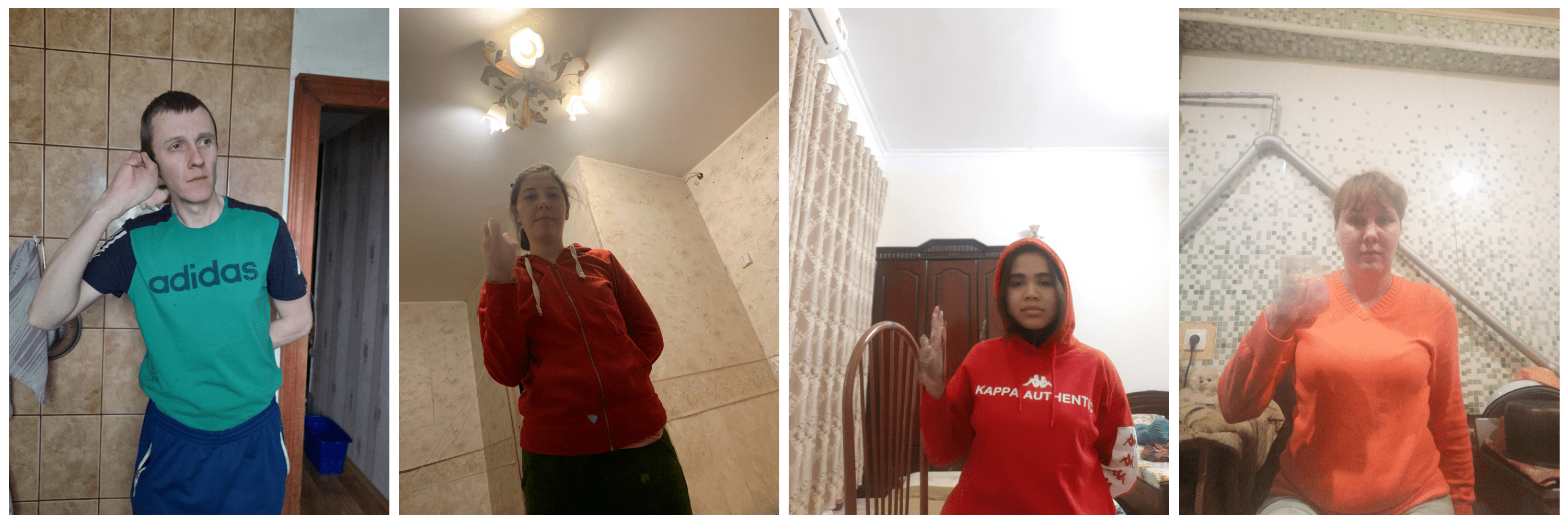}
    \caption{Redrawn samples by Stable Diffusion.}
    \label{fig:redrawn}
\end{figure*}

Our approach began by randomly selecting 30,000 images from the HaGRID dataset. Since these images are authentic photographs, all hands within this chosen subset are classified as standard hands. We employed the Stable Diffusion model \cite{2022_ldm} to recreate the hand areas, as outlined by their bounding boxes in HaGRID images. Bounding box information was directly obtained from the HaGRID dataset. Consequently, for every original image from HaGRID, we created a corresponding redrawn image. Examples of these redrawn images are displayed in \autoref{fig:redrawn}.

Due to certain limitations of the Stable Diffusion model, some hands in the redrawn images were classified as non-standard hands. From these redrawn images, we manually selected samples featuring non-standard hands, pairing each with its corresponding original image, which depicts a standard hand. We annotated each image in this set with labels and bounding boxes to highlight the presence and location of the hands. To enable a comprehensive evaluation of our model, we divided this data into training and testing sets. This dataset is specifically prepared for detecting non-standard hands.

\textbf{Hand Detection}

% YOLOv8 \cite{yolov8} is a state-of-the-art model for object detection, instance segmentation, image classification, and pose estimation. To locate the hand and classify as non-standard hand or standard hand, we use training set mentioned above to fine-tune the YOLOv8 model. 

\begin{figure*}[!bt]
    \centering
    \includegraphics[height=16.8cm]{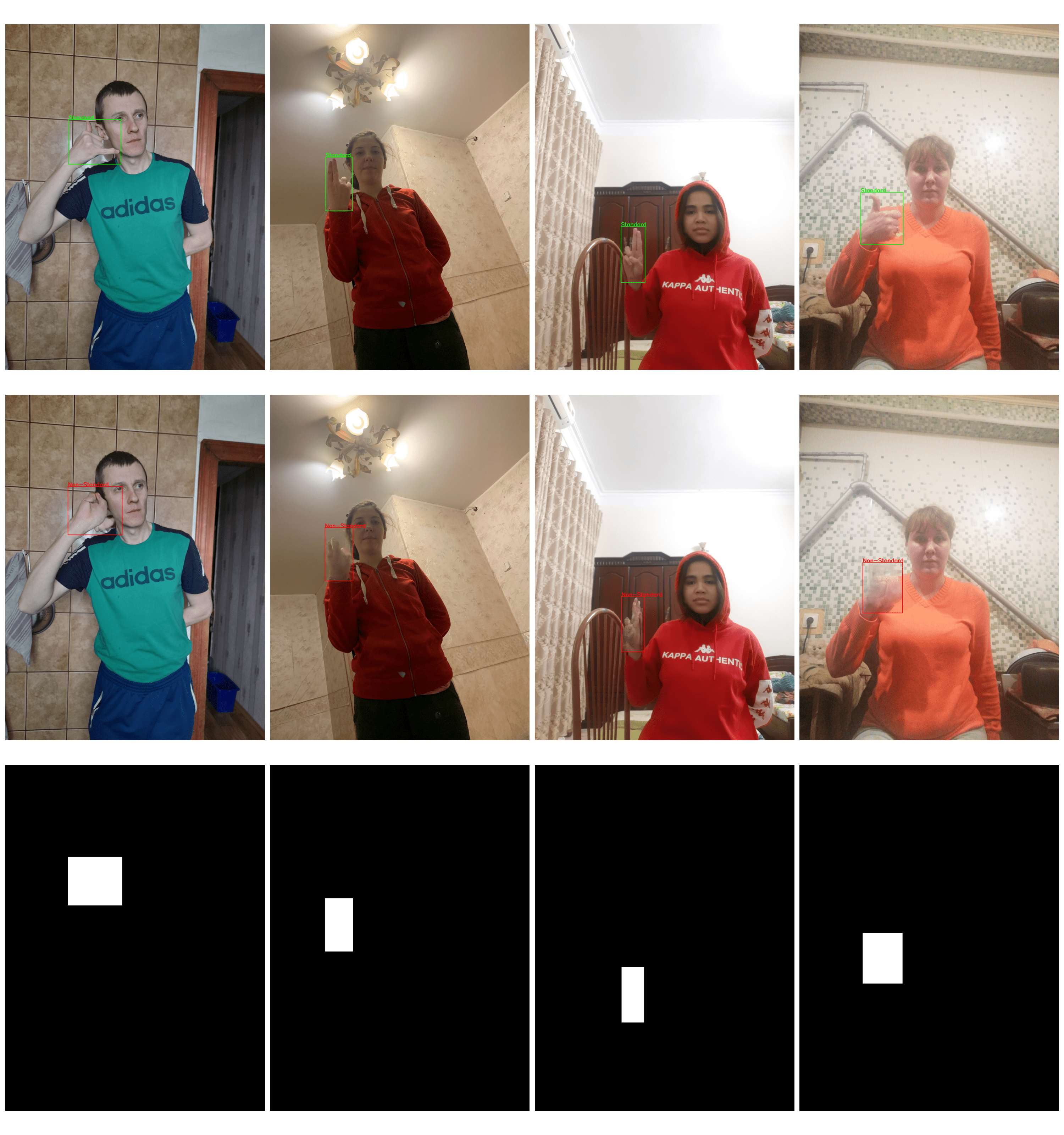}
    \caption{Original samples with detected bounding box, redrawn samples with detected bounding box and redrawn samples with bounding box mask.}
    \label{fig:bbox}
\end{figure*}

YOLOv8 \cite{yolov8} is one of the state-of-the-art models for object detection. To identify and classify hands as either non-standard or standard, we finetuned the YOLOv8 model using the training dataset described above. After implementing the trained YOLO model, we annotate the bounding boxes around the hands and classify them. Only the non-standard hands are chosen for further restoration. These bounding boxes are transformed into the 'bounding box masks', as shown in \autoref{fig:bbox}. All examples are derived from the test set. The illustration demonstrates the finetuned model's ability to accurately locate and classify the hand.

% Upon deploying the trained YOLO model, the bounding boxes of hands are annotated, and the hands are classified accordingly. Only the non-standard hands are retained for restoration. The bounding boxes of these hands are converted into masks, which we refer to as bounding box mask. For a clearer understanding, we provide visual illustration in \autoref{fig:bbox}. All samples are selected from test set. From illustration we can know that fine-tuned model can correctly locate and classify the hand.

\subsection{Body Pose Estimation}

\begin{figure*}[!bt]
    \centering
    \includegraphics[width=\textwidth]{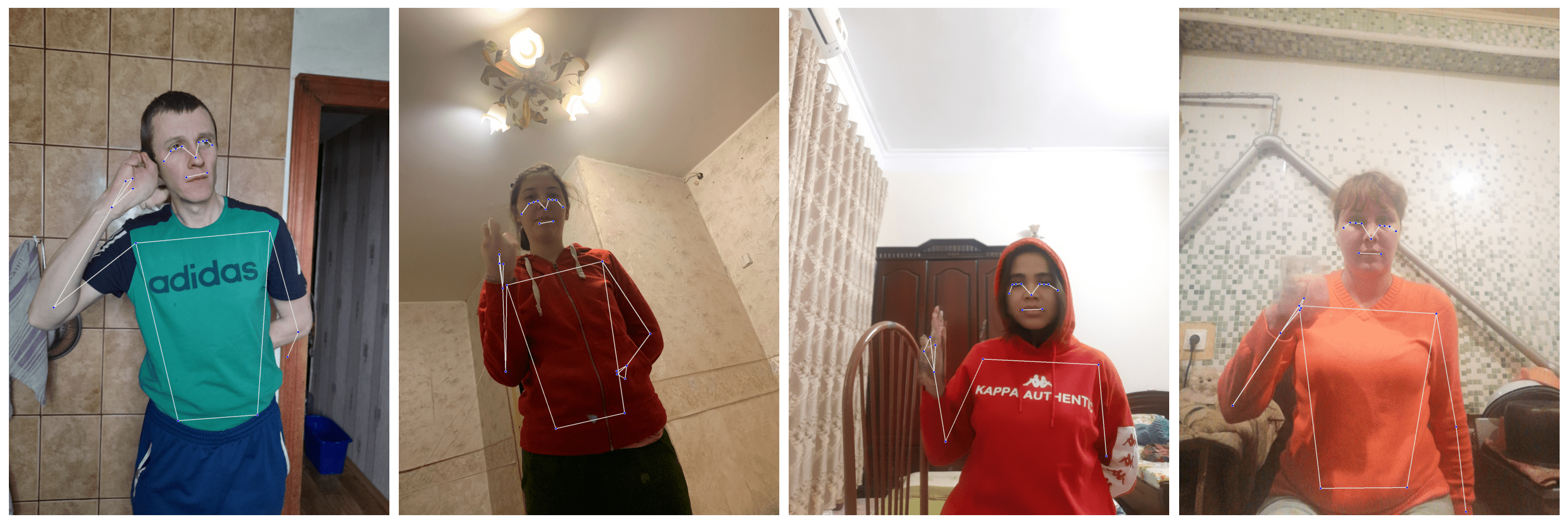}
    \caption{Redrawn samples with body skeleton detected by MediaPipe.}
    \label{fig:mediapipe}
\end{figure*}

Estimating body pose is crucial for determining the size position, and chirality of hands in our images. MediaPipe~\cite{mediapipe} provides various solutions for vision tasks. Notably, its pose landmark detection feature is capable of detecting the human body skeleton. This includes a machine learning model skilled at identifying body landmarks such as hands, elbows, shoulders, hips, and more in images or videos, and their structural interconnections, as shown in \autoref{fig:mediapipe}.

\begin{figure*}[!ht]
	\centering
	\begin{subfigure}[b]{0.24\textwidth}
		\centering
		\includegraphics[width=\linewidth, keepaspectratio]{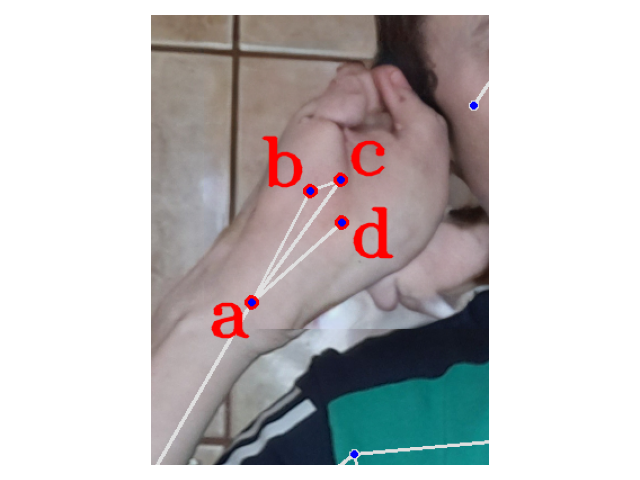}
		\caption{Hand skeleton.}
		\label{fig:hand_skeleton}
	\end{subfigure}
	\hfill % add some space between figures
	\begin{subfigure}[b]{0.24\textwidth}
		\centering
		\includegraphics[width=\linewidth, keepaspectratio]{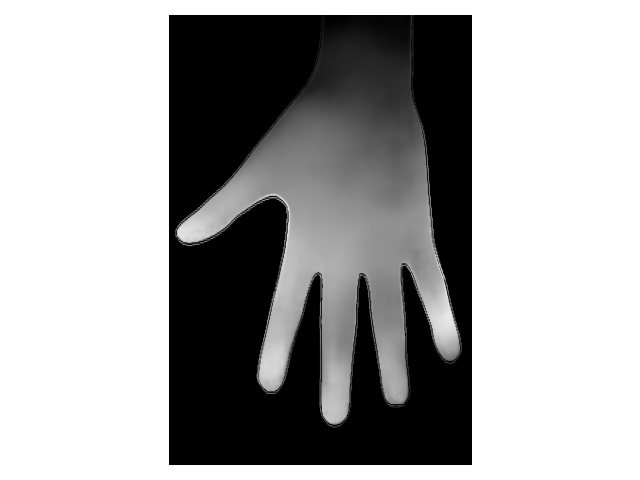}
		\caption{Template: opened-palm.}
		\label{fig:template-a}
	\end{subfigure}
	\hfill % add some space between figures
        \begin{subfigure}[b]{0.24\textwidth}
		\centering
		\includegraphics[width=\linewidth, keepaspectratio]{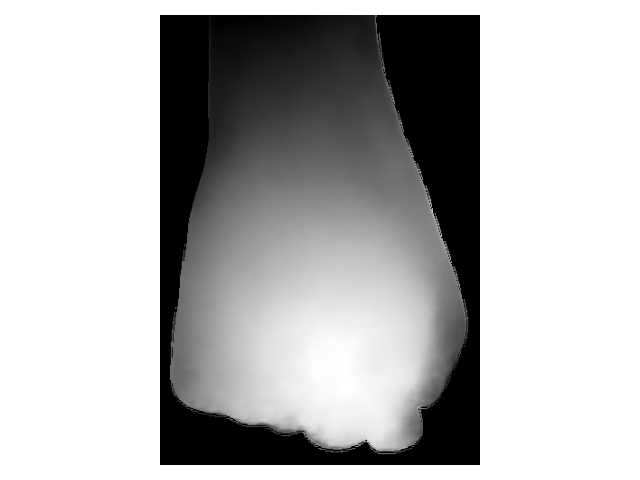}
		\caption{Template: fist-back.}
		\label{fig:template-b}
	\end{subfigure}
	\hfill % add some space between figures
	\begin{subfigure}[b]{0.24\textwidth}
		\centering
		\includegraphics[width=\linewidth, keepaspectratio]{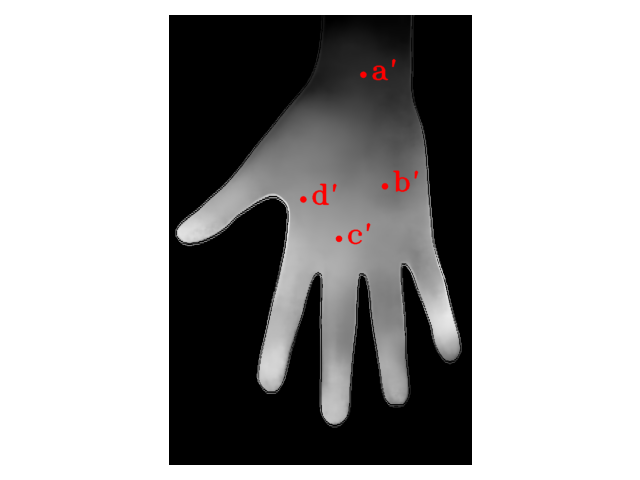}
		\caption{Template with landmark.}
		\label{fig:template_skeleton}
	\end{subfigure}
	\caption{Hand and template details.}
	\label{fig:hand_and_template_details}
\end{figure*}

We employ MediaPipe because it provides three unique landmarks for the hand, as depicted in \autoref{fig:hand_skeleton}. These landmarks are crucial for accurately determining the hand's size, position, and gesture chirality. In comparison, other 2D pose estimation models \cite{alphapose, xu2022vitpose} typically provide information only up to the wrist. Moreover, MediaPipe's detailed body skeleton detection remains reliable even when hand images in non-standard hand pictures are blurry or unclear, ensuring a certain degree of prediction accuracy.

% Upon detection, the skeletal landmarks were superimposed onto the redrawn image. For a more illustrative representation, the hand region were extracted and showcased in \autoref{fig:hand_skeleton}. The hand presents four salient landmarks: $a$, $b$, $c$, and $d$. Defining two vectors, $v_1 = \overrightarrow{ac}$ and $v_2 = \overrightarrow{bd}$, we compute their cross product $v_1 \times v_2$. The sign of the resultant aids in the classification of the hand orientation. If the result is negative, the hand is classified as a \textit{CW hand} (indicating a clockwise rotation from $v_1$ to $v_2$). Conversely, a positive result classifies the hand as a \textit{CCW hand} (signifying a counter-clockwise rotation from $v_1$ to $v_2$). This methodological classification offers us a straightforward approach to discern the chirality of the hand.

Once detected, the skeletal landmarks were overlaid on the redrawn image. The hand region, highlighting these landmarks, is exhibited in \autoref{fig:hand_skeleton}. Four key landmarks are identified on the hand: $a$, $b$, $c$, and $d$. By defining vectors $v_1 = \overrightarrow{ac}$ and $v_2 = \overrightarrow{bd}$, we calculate their cross product $v_1 \times v_2$. The sign of this cross product helps classify hand orientation. A negative value indicates a \textit{CW hand} (clockwise rotation from $v_1$ to $v_2$), while a positive value signals a \textit{CCW hand} (counter-clockwise rotation from $v_1$ to $v_2$). This method simplifies the process of determining the hand's chirality.

\subsection{Control Image Generation}

\begin{figure*}[!ht]
	\centering
	\begin{subfigure}[b]{0.3\textwidth}
		\centering
		\includegraphics[width=\linewidth, keepaspectratio]{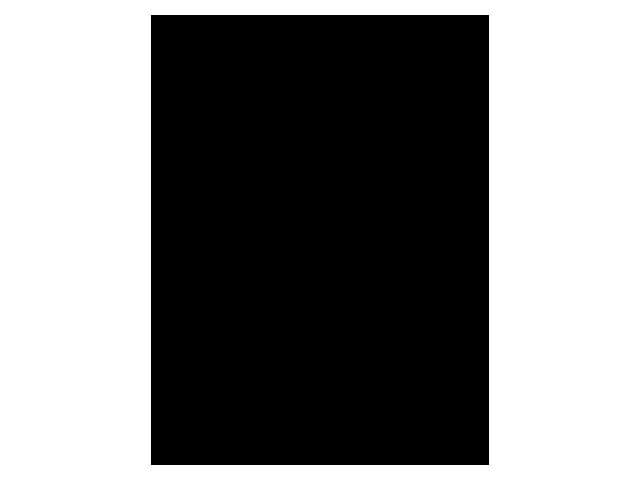}
		\caption{Background.}
		\label{fig:background}
	\end{subfigure}
	\hfill % add some space between figures
	\begin{subfigure}[b]{0.3\textwidth}
		\centering
		\includegraphics[width=\linewidth, keepaspectratio]{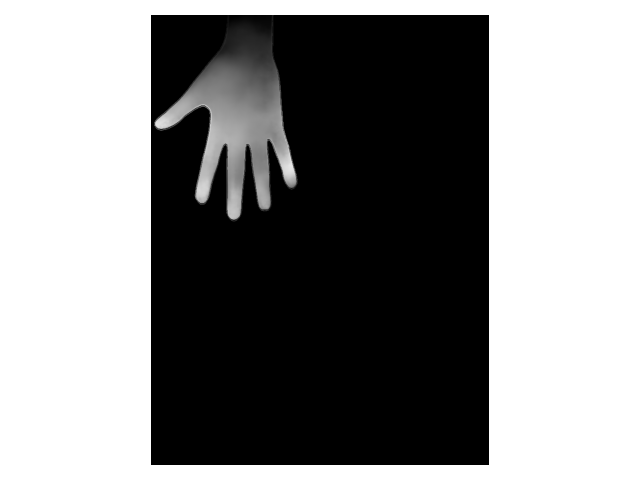}
		\caption{Template in background.}
		\label{fig:template-in-background}
	\end{subfigure}
	\hfill % add some space between figures
	\begin{subfigure}[b]{0.3\textwidth}
		\centering
		\includegraphics[width=\linewidth, keepaspectratio]{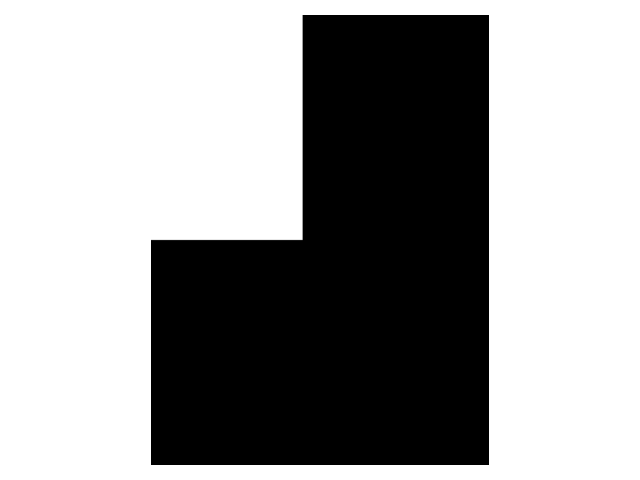}
		\caption{Template mask in background.}
		\label{fig:template-mask-in-background}
	\end{subfigure}
	\caption{Background, template in background, and template mask in background.}
\end{figure*}

\begin{figure*}[!ht]
	\centering
	\begin{subfigure}[b]{\textwidth}
		\centering
		\includegraphics[height=5cm, keepaspectratio]{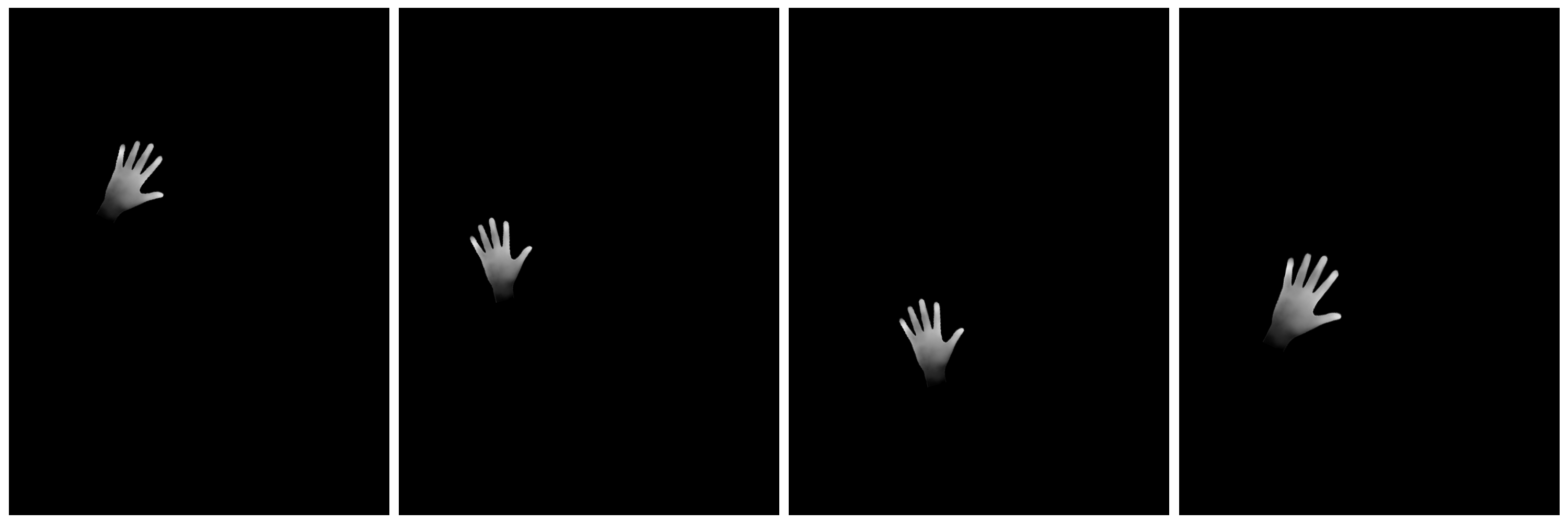}
		\caption{Control image of redrawn samples.}
		\label{fig:control-image}
	\end{subfigure}

	\begin{subfigure}[b]{\textwidth}
		\centering
		\includegraphics[height=5cm, keepaspectratio]{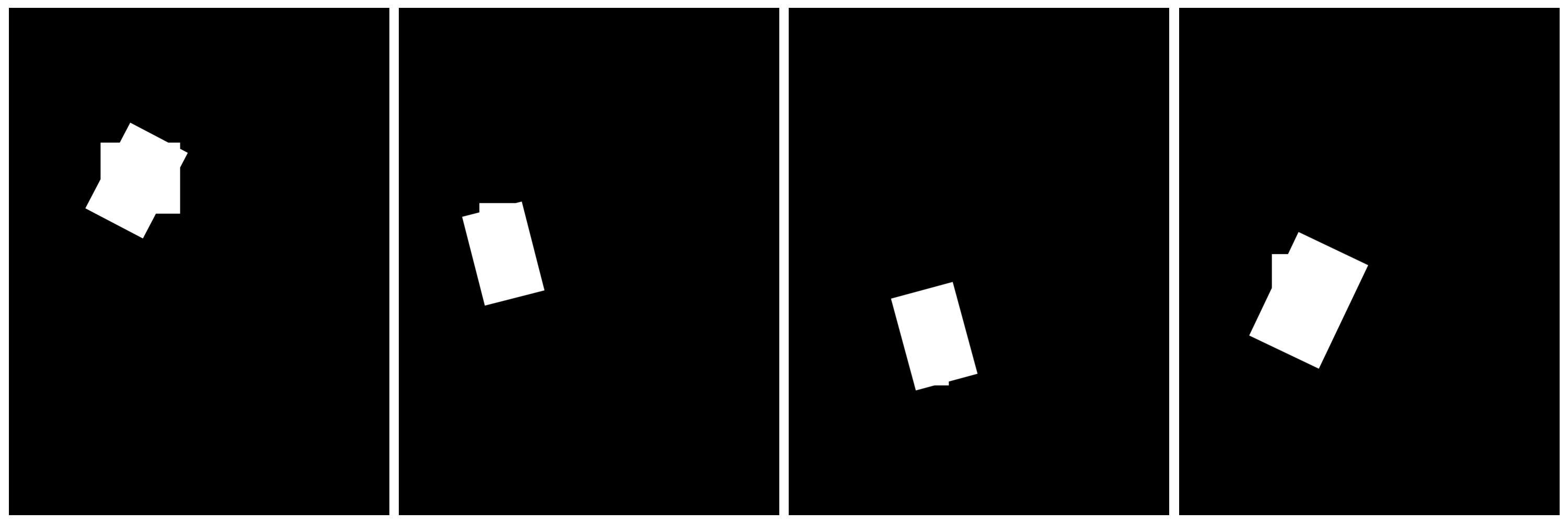}
		\caption{Union mask of redrawn samples.}
		\label{fig:union-mask}
	\end{subfigure}

	\begin{subfigure}[b]{\textwidth}
		\centering
		\includegraphics[height=5cm, keepaspectratio]{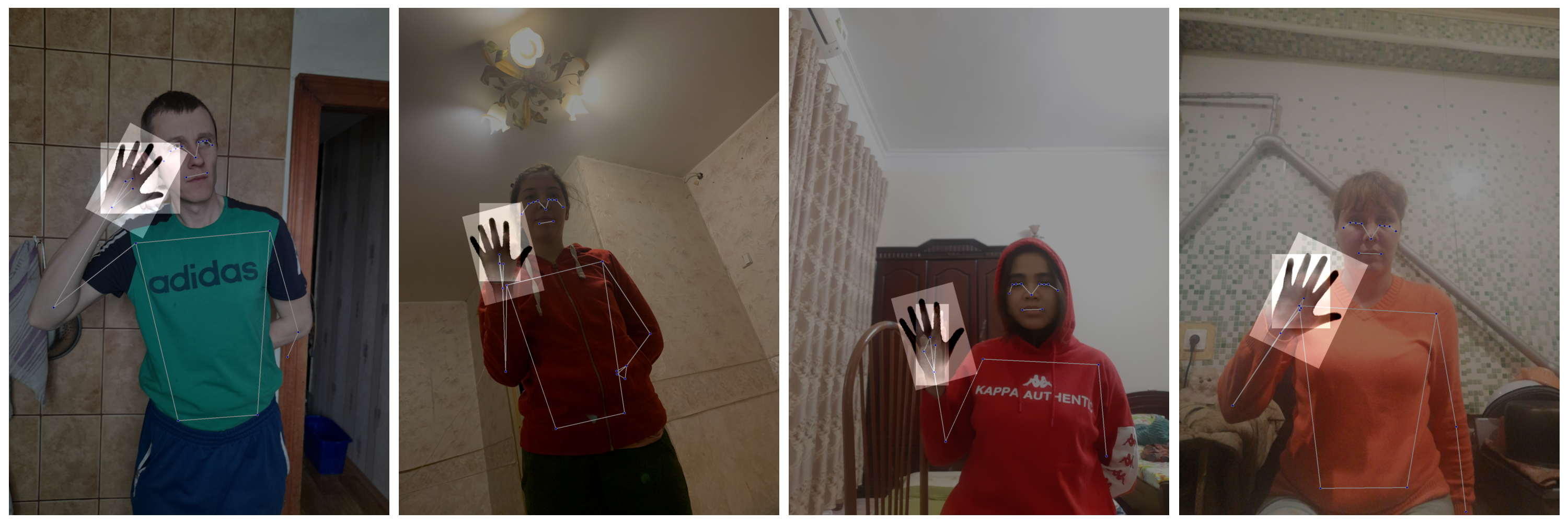}
		\caption{Visualization of redrawn samples, body skeleton, control image, and union mask.}
		\label{fig:visualization}
	\end{subfigure}
	\caption{Some samples in control image generation.}
	\label{fig:samples-control}
\end{figure*}

% ControlNet \cite{2023_controlnet} has emerged as an enhancement to the Stable Diffusion model. By incorporating additional input modalities, such as key landmarks, depth maps, edge maps, among others, ControlNet steers Stable Diffusion towards generating images with a more stable and realistic composition. This is important when aiming to restore hand representations in images, ensuring that the resultant shape, size, and contours adhere closely to real-world standards.

ControlNet \cite{2023_controlnet} enhances the Stable Diffusion model by introducing additional input modalities, including key landmarks, depth maps, and edge maps. This integration directs Stable Diffusion to produce images with more stable and realistic structures. Such advancements are crucial in restoring hand images, ensuring that their shape, size, and contours align accurately with real-world characteristics.

In refining the restoration process, we introduce a ControlNet image as an additional conditioning input. This image, tailored to enhance Stable Diffusion's restoration of non-standard hands, pushes the model to its limits and improves the output quality. Examples~\cite{kamph2023youtube} include opened-palm (see \autoref{fig:template-a}) and fist-back (see \autoref{fig:template-b}), collectively termed as hand templates. Adding more templates is feasible. A key part of this process is accurately placing these hand-templates at the correct hand locations, forming the Control Image. The accuracy of this placement is crucial as it significantly impacts ControlNet's effectiveness in rendering a precise hand representation. Our methodology for this includes:

\begin{enumerate}
    \item Like the hand landmarks identified by MediaPipe, we annotate four specific points ($a'$, $b'$, $c'$, $d'$) on the Templates, as shown in \autoref{fig:template_skeleton}. This helps us form vectors $v_1' = \overrightarrow{a'c'}$ and $v_2' = \overrightarrow{b'd'}$.
    
    \item We select an appropriate Template based on the gesture and context of the image undergoing restoration.
    
    \item A background image, identical in size to the redrawn image but entirely black, is prepared (see \autoref{fig:background}). The Template is initially positioned at the top-left corner of this background (see \autoref{fig:template-in-background}). Simultaneously, a white image of the same size as the Template, called the template mask, is also placed at the top-left of the background (see \autoref{fig:template-mask-in-background}).
    
    \item The chirality (CW or CCW) of the hand is determined. If it does not match the Template's chirality, the template is flipped.
    
    \item \textbf{Scaling}: The Template and template mask are scaled by the ratio $\frac{|v_1|}{|v_1'|}$.
    
    \item \textbf{Moving}: The Template and template mask are moved by the vector difference $v_1 - v_1'$.
    
    \item \textbf{Rotation}: The Template and template mask are rotated by an angle $\theta$. The angle is calculated as:
    
    $$\theta = \arccos{\frac{a a' + c c'}{\sqrt{a^2+c^2}\sqrt{a'^2+c'^2}}}$$
    
    If $a \times c' - c \times a' > 0$, we rotate clockwise; otherwise, we rotate counterclockwise.
    
    \item The final image with the Template is termed the control image (see \autoref{fig:control-image}).
    
    \item The combination of the template mask and the bounding box mask forms the union mask (see \autoref{fig:union-mask}).
    
    \item For a comprehensive view of the process, we compile samples of the redrawn images, body skeleton, control image, and union mask into a single visual representation (see \autoref{fig:visualization}).
\end{enumerate}

% It is necessary to use both YOLOv8 and MediaPipe results for this step, because MediaPipe can detect the occluded hand and cannot distinguish whether it is a non-standard hand, and YOLOv8 can compensate for this deficiency. By above methodology, we ensure that the Control Image provides the optimal cues to guide the ControlNet in generating accurate hand reconstructions.

Utilizing both YOLOv8 and MediaPipe results is crucial for this step. While MediaPipe detects occluded hands, it cannot differentiate between non-standard and standard hands. YOLOv8 compensates for this limitation. Through the methodology described above, we guarantee that the Control Image offers essential cues for ControlNet to accurately reconstruct hand images.

\subsection{ControlNet Inpainting}

\begin{figure*}[!h]
\centering
\includegraphics[width=1\textwidth]{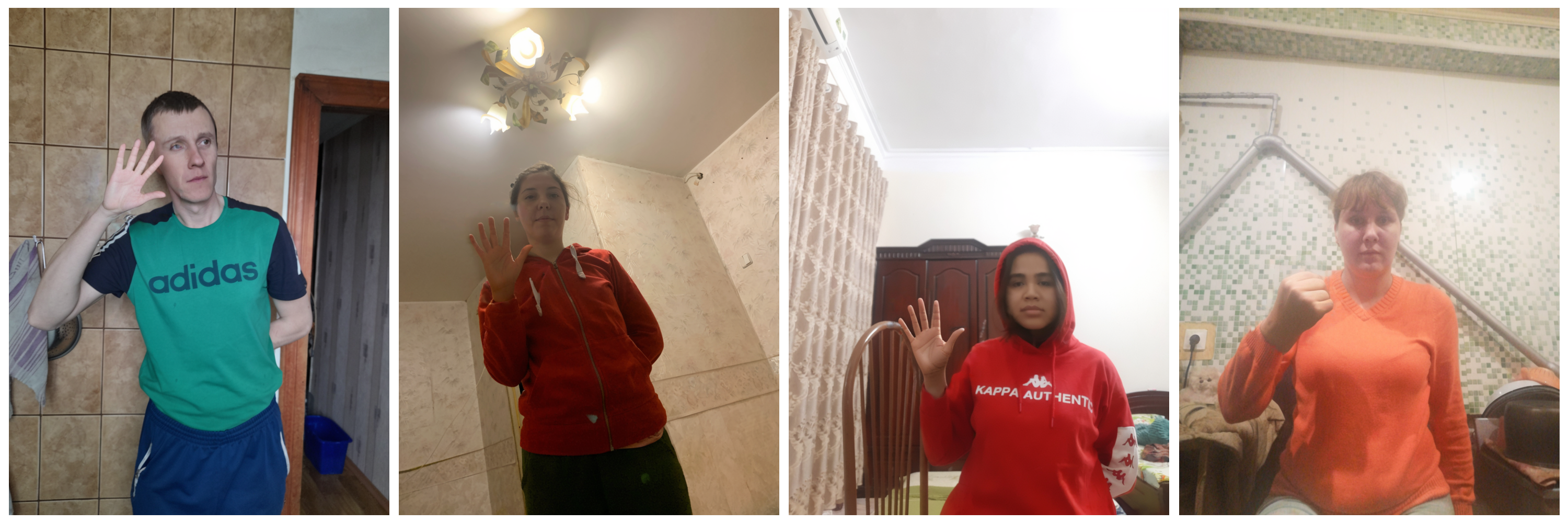}
\caption{ControlNet inpainting result.}
\label{fig:controlnet}
\end{figure*}

After establishing the hand's position and shape, we engage ControlNet for image restoration. This advanced process relies on the Union Mask to focus restoration efforts on specific areas of the image for precise improvement.

The control image, accurately depicting the hand's position, acts as a guide for ControlNet in restoring the redrawn image. We enhance this process with carefully selected prompts, aimed at providing detailed cues for desired image attributes. Our primary prompt is: 
\begin{quote}
    \textit{[TEMPLATE NAME]}, hand, realskin, photorealistic, RAW photo, best quality, realistic, photo-realistic, masterpiece, an extremely delicate and beautiful, extremely detailed, 2k wallpaper, Amazing, finely detailed, 8k wallpaper, huge filesize, ultra-detailed, high-res, and extremely detailed.
\end{quote}

We also use `negative prompts' to avoid unwanted outcomes, including: ``deformed, EasyNegative, paintings, sketches, (worst quality:2), (low quality:2), (normal quality:2), low-res, normal quality, and (monochrome).''

These prompts, familiar within the community for generating high-quality images, reflect prompt engineering domain knowledge. Using both positive and negative prompts allows us to fully utilize ControlNet, producing images that are not only aesthetically pleasing but also realistic, as shown in \autoref{fig:controlnet}.

\subsection{IP2P Inpainting}

\begin{figure*}[!h]
\centering
\includegraphics[width=1\textwidth]{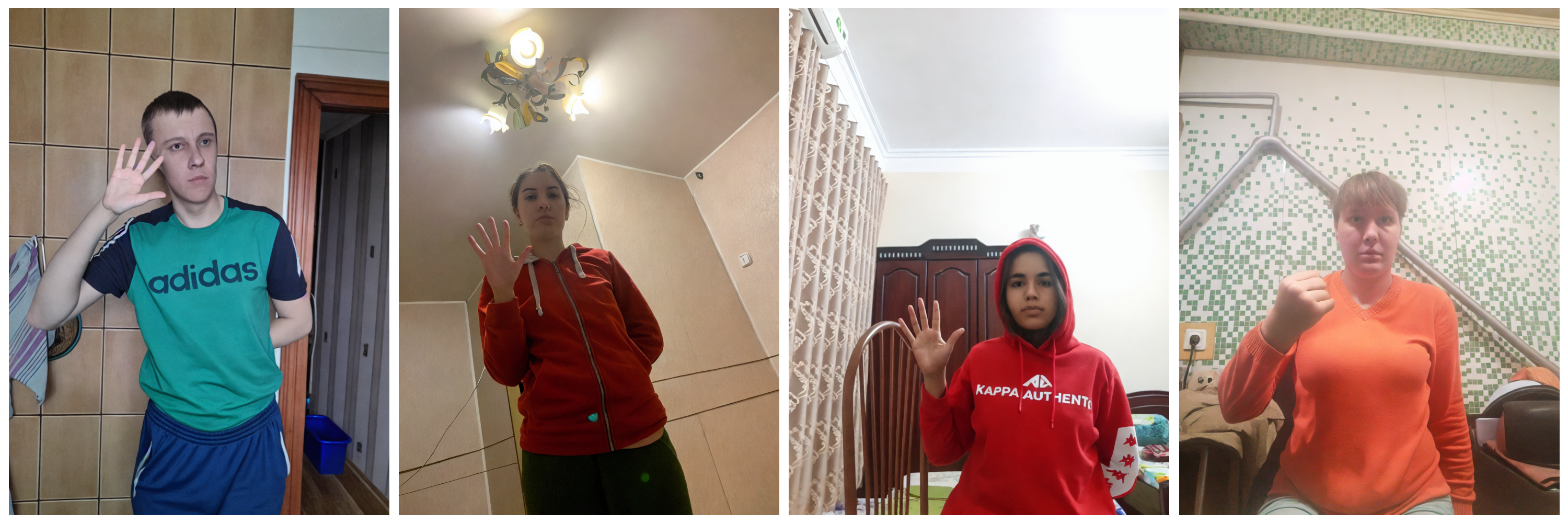}
\caption{IP2P inpainting result.}
\label{fig:ip2p}
\end{figure*}

The final phase of our method employs the \textit{InstructPix2Pix} (IP2P) model \cite{brooks2022instructpix2pix}. Following the initial restoration with ControlNet, IP2P enhances the textures, focusing on giving the hands a more realistic and authentic appearance to blend seamlessly with the rest of the image.

% Training Part
Initially, the IP2P model is fine-tuned using our training set, which comprises 9623 pairs of images. Each pair includes a real image with a standard hand from the HaGRID dataset and a corresponding Stable Diffusion redrawn image with a non-standard hand. The real images were inputs during training, while the redrawn images served as outputs. Additionally, a prompt, ``Turn the deformed hand into normal'', and its 50 variants (see \autoref{chap:appendix1}) were also used as input. This fine-tuning enables the IP2P model to transform non-standard hands into standard ones with enhanced accuracy.

For restoration, the fine-tuned model is then applied. Unlike previous steps, no masking is used here. The entire image undergoes processing to ensure the hand's texture matches the overall image style. This comprehensive approach employs the prompt: ``Turn the deformed hand into normal''.

The outcome of this process, the final result of our restoration effort, is displayed in \autoref{fig:ip2p}.

\section{Conclusion}

Through the application of our approach, we observed encouraging results. The detection phase, utilizing the YOLO model \cite{yolov8}, accurately pinpointed hand locations in images and classified them as non-standard or standard hands. We evaluated a test set of 2006 images, representing diverse scenes and gestures, using Precision and Recall metrics of 0.85, 0.90, 0.95. Our models performed well across these indicators, demonstrating the effectiveness and adaptability of our method in various real-world scenarios. This includes effectively detecting non-standard hands in images generated by Stable Diffusion.

The restoration phase further strengthened these outcomes. Each step, from body pose estimation to control image creation and inpainting processes, was precisely executed. We demonstrated using FID that the outcomes of ControlNet inpainting improved upon images with non-standard hands, and that IP2P inpainting further enhanced the results from ControlNet. Both ControlNet and IP2P inpainting proved effective. We tested not only with redrawn samples from the HaGRID dataset but also with images generated by Stable Diffusion and real photographs. Our experiments show non-standard hand images transforming to more closely resemble standard hands.

\newpage
\bibliography{bib}
\bibliographystyle{icml2023}

%%%%%%%%%%%%%%%%%%%%%%%%%%%%%%%%%%%%%%%%%%%%%%%%%%%%%%%%%%%%%%%%%%%%%%%%%%%%%%%
%%%%%%%%%%%%%%%%%%%%%%%%%%%%%%%%%%%%%%%%%%%%%%%%%%%%%%%%%%%%%%%%%%%%%%%%%%%%%%%
% APPENDIX
%%%%%%%%%%%%%%%%%%%%%%%%%%%%%%%%%%%%%%%%%%%%%%%%%%%%%%%%%%%%%%%%%%%%%%%%%%%%%%%
%%%%%%%%%%%%%%%%%%%%%%%%%%%%%%%%%%%%%%%%%%%%%%%%%%%%%%%%%%%%%%%%%%%%%%%%%%%%%%%
% \newpage
% \appendix
% \onecolumn

% You can have as much text here as you want. The main body must be at most $8$ pages long.
% For the final version, one more page can be added.
% If you want, you can use an appendix like this one, even using the one-column format.
%%%%%%%%%%%%%%%%%%%%%%%%%%%%%%%%%%%%%%%%%%%%%%%%%%%%%%%%%%%%%%%%%%%%%%%%%%%%%%%
%%%%%%%%%%%%%%%%%%%%%%%%%%%%%%%%%%%%%%%%%%%%%%%%%%%%%%%%%%%%%%%%%%%%%%%%%%%%%%%

\end{document}